\documentclass{article}
\usepackage{arxiv}

\usepackage{amsmath,amsfonts,bm}

\def\eqref#1{equation~\ref{#1}}

\def\1{\bm{1}}

\DeclareMathAlphabet{\mathsfit}{\encodingdefault}{\sfdefault}{m}{sl}
\SetMathAlphabet{\mathsfit}{bold}{\encodingdefault}{\sfdefault}{bx}{n}

\usepackage{xcolor}
\definecolor{lightblue}{rgb}{0.6, 0.8, 0.9}
\definecolor{darkblue}{rgb}{0.2,0.4,0.6}
\definecolor{darkgreen}{rgb}{0, 0.55, 0.12}
\definecolor{darkred}{rgb}{0.6,0,0}
\usepackage[
  breaklinks=true,
  colorlinks,
  linkcolor=darkblue,
  citecolor=darkblue,
  bookmarks=false
]{hyperref}
\usepackage{url}

\usepackage{tikz}
\usepackage{graphicx}
\usepackage{wrapfig}
\usepackage{booktabs}
\usepackage{adjustbox}
\usepackage{multirow}
\usepackage{xspace}
\usepackage{array}
\usepackage{bm}
\usepackage{bbm}
\usepackage{amsthm, amsmath, amssymb}
\usepackage{cleveref}
\usepackage{algorithm}
\usepackage{algorithmic}

\usepackage{listings}
\usepackage{wrapfig}
\usepackage{caption}
\usepackage{subcaption}
\usepackage{url}
\usepackage{colortbl}
\usepackage{adjustbox}
\usepackage{makecell}
\usepackage{pifont}
\usepackage{tcolorbox}
\usepackage{setspace}
\usepackage{fancybox}
\usepackage{tocloft}
\usepackage{etoc}
\usepackage{enumitem}
\usepackage{pifont}
\usepackage{bm}
\usepackage{ulem}
\definecolor{mymauve}{rgb}{0.58,0,0.82}
\newlength{\mysize}

\allowdisplaybreaks

\newtheorem{thm}{Theorem}
\newtheorem{lem}[]{Lemma}

\theoremstyle{definition}

\newtheorem{assumption}[thm]{Assumption}

\title{Principled Multimodal Representation Learning}

\author{Xiaohao Liu \ \ \ \ \ Xiaobo Xia\thanks{Corresponding author.} \ \ \ \ \ See-Kiong Ng \ \ \ \ \ Tat-Seng Chua \\
National University of Singapore \\
\tt\small xiaohao.liu@u.nus.edu \quad xiaoboxia.uni@gmail.com \quad 
seekiong@nus.edu.sg \quad dcscts@nus.edu.sg
}

\begin{document}

\maketitle

\begin{abstract}
Multimodal representation learning seeks to create a unified representation space by integrating diverse data modalities to improve multimodal understanding. Traditional methods often depend on pairwise contrastive learning, which relies on a predefined anchor modality, restricting alignment across all modalities. Recent advances have investigated the simultaneous alignment of multiple modalities, yet several challenges remain, such as limitations imposed by fixed anchor points and instability arising from optimizing the product of singular values. To address the challenges, in this paper, we propose Principled Multimodal Representation Learning (PMRL), a novel framework that achieves simultaneous alignment of multiple modalities without anchor dependency in a more stable manner. Specifically, grounded in the theoretical insight that full alignment corresponds to a rank-1 Gram matrix, PMRL optimizes the dominant singular value of the representation matrix to align modalities along a shared leading direction. We propose a softmax-based loss function that treats singular values as logits to prioritize the largest singular value. Besides, instance-wise contrastive regularization on the leading eigenvectors maintains inter-instance separability and prevents representation collapse. Extensive experiments across diverse tasks demonstrate PMRL’s superiority compared to baseline methods. Source code can be found in \url{https://github.com/Xiaohao-Liu/PMRL}.
\end{abstract}

\addtocontents{toc}{\protect\setcounter{tocdepth}{-1}}

\section{Introduction}

Humans perceive the world through a rich interplay of multimodal signals, integrating visual, auditory, textual, and tactile information to form cohesive representations of individual instances~\cite{ngiam2011multimodal,lu2023theory, zong2024self,  xu2023multimodal, zhu2024vision,luo2024mmevol,shi2023dynamic,baltruvsaitis2018multimodal}. These modalities capture both shared and distinct concepts, completing one instance while enabling the differentiation from another. Inspired by this capability, multimodal representation learning~(MRL) seeks to align diverse modalities within a unified space~\cite{wu2022wav2clip, radford2021learning, girdhar2023imagebind, zhu2023languagebind, wang2024omnibind, liu2025continual, dufumier2024align}, where a representation from one modality can effectively retrieve or reconstruct corresponding representations from others.

\begin{figure*}[t]
    \centering
    \includegraphics[width=0.96\linewidth]{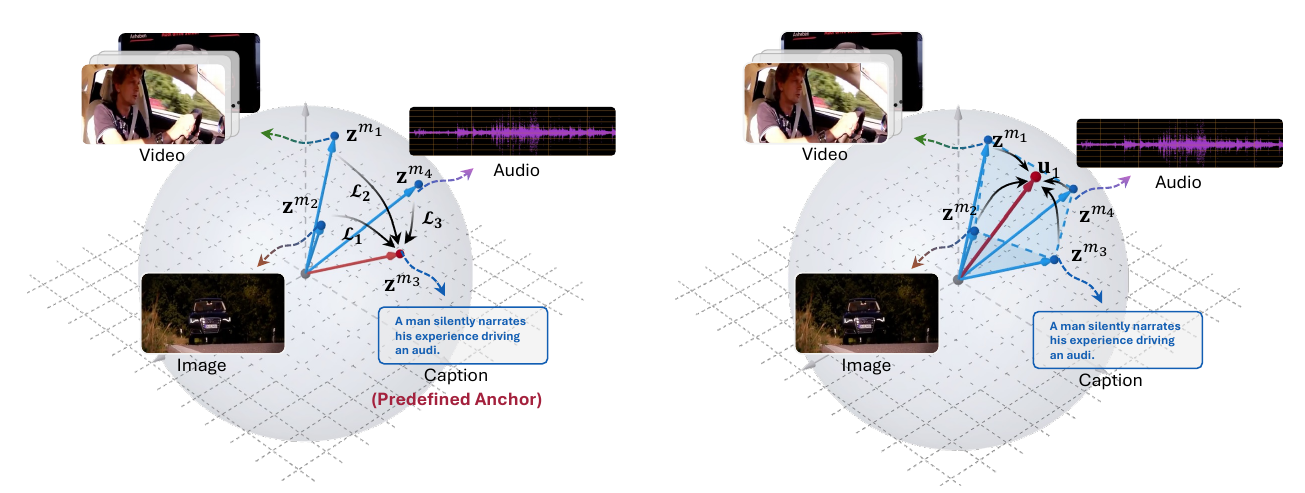}
    \caption{\textbf{The illustration of multimodal representations within a hypersphere.} The left demonstrates pairwise contrastive learning to align multiple modalities with a predefined anchor (\textit{i.e.}, caption), where modalities are sampled to multiple pairs. The right illustrates our method that aligns all modalities simultaneously with a leading direction.}
    \label{fig:top}
\end{figure*}

Bimodal alignment, often achieved with contrastive learning~\cite{he2020momentum, chen2020simple, wang2020understanding}, aligns one modality with another by comparing synthetic modality pairs~\cite{yang2020learning, xu2021videoclip, wu2022wav2clip, zhang2022pointclip}. This paradigm demonstrates remarkable performance in tasks like image-text or audio-text understanding. Such success can also be replicated in MRL. Modality-binding methods, exemplified by ImageBind~\cite{girdhar2023imagebind}, designate one modal anchor as the centroid and adopt pairwise contrastive learning to align other modalities to it~\cite{zhu2023languagebind,lyu2024unibind,guo2023point, wang2024omnibind, wang2024freebind}, as shown in Figure~\ref{fig:top}(left). Anchored alignment (\textit{e.g.}, $m_1\rightarrow m_3$ and $m_4\rightarrow m_3$ ) is explicitly modeled, while the alignment among non-anchor modalities (\textit{e.g.}, $m_1\rightarrow m_4$) remains implicit. 
A branch of work proposes exploiting scaling data for pre-training~\cite{bain2021frozen, nagrani2022learning, wang2023internvid, zhu2023languagebind, chen2023vast, zhao2024videoprism, lyu2024unibind}, or introducing auxiliary learning objectives, like language modeling loss~\cite{li2022blip, wang2022omnivl, chen2023vast} to improve MRL. 
Unfortunately, they remain reliant on pairwise contrastive learning, which keeps the alignment hinged with anchors.

A most recent work attempts to move beyond this holding paradigm by minimizing the volume of a parallelotope formed by multimodal representations~\cite{cicchetti2024gramian}. It utilizes the determinant of the Gram matrix (numerically equal to the product of singular values) and interprets simultaneous alignment for all modalities in a geometric space. 
Unfortunately, it still depends on \textit{predefined anchors} to construct negative instances (\textit{i.e.}, replace the content of anchor modality to yield an unmatched multimodal instance). Moreover, its optimization on volume suffers from \textit{instability}. Specifically, when the parallelotope collapses to a plane,
optimization halts as the volume reaches zero, resulting in incomplete alignment. We also discuss this via singular value analysis~(see Section~\ref{sec:further_analysis}).
These limitations underscore the need for a more advanced MRL method with sound principles, motivating our framework development for multimodal alignment.

In this work, we initiate our research from the foundational goal of multimodal alignment, aiming to maximize the similarity between any modality pairs of a shared instance. This leads to a critical insight: establishing a fundamental connection between multimodal alignment and the rank of the Gram matrix, where full alignment is achieved when the rank equals one. This principle guides our development of a novel method for multimodal representation learning through rank-1 matrix approximation.
To advance this, we propose strengthening the maximum singular value to encourage full alignment, drawing on the optimal low-rank approximation theory. 
The maximum singular value corresponds to a leading direction (\textit{i.e.}, the dominant eigenvector), specialized for different instances. As this singular value increases, multimodal representations are aligned toward this direction adaptively, as depicted in Figure~\ref{fig:top}(right). 
This leading direction drifts with data itself, rather than privileging any one of the modalities.
Motivated and implemented by the principle, we term our method as Principled Multimodal Representation Learning (PMRL) to highlight its theoretical grounding and pioneering design. 
PMRL removes anchor constraints, elevating any-to-one alignment to a straightforward \textit{any-to-any} alignment for MRL. In addition, optimizing the maximum singular value relative to their sum also ensures \textit{greater stability} than the previous volume-based method.

To this end, we propose a novel learning objective that directly aligns all modalities by optimizing singular values. Specifically, we employ a softmax-based loss that treats the singular values as logits and emphasizes the dominance of the maximum singular value. Besides, we incorporate instance-wise contrastive regularization over leading eigenvectors. These vectors serve as alignment centroids and are regularized to ensure inter-instance separability and prevent representation collapse.  We verify the proposed PMRL with extensive experiments and demonstrate its superiority compared to baselines. 

Before delving into details, we summarize our contributions as follows:
\begin{itemize}
    \item We introduce Principled Multimodal Representation Learning (PMRL), a novel framework that encourages full alignment across multiple modalities simultaneously without relying on a predefined anchor modality. Our method is grounded in a theoretical connection between the singular values of multimodal representations and their full alignment under rank-1 approximation. 
    \item To operationalize this insight, we reformulate the learning objective to strengthen the dominance of the maximum singular value, promoting full alignment across modalities, and incorporate instance-wise regularization to enhance inter-instance separability. By optimizing the maximum singular value, our method stabilizes the learning compared to directly reducing the products of singular values (\textit{i.e.}, determinant-based volume).
    \item Extensive experiments on diverse tasks, including text-video retrieval, text-audio retrieval, and downstream classification, demonstrate PMRL’s superior performance. Note that PMRL is capable of enhancing representations for broader fields, like medical applications (\textit{e.g.}, autism diagnosis). Comprehensive analyses, including ablation studies, singular value analysis, regularization effects, noise robustness, and modality contribution, validate the efficacy and design rationale of PMRL, establishing its potential for advancing multimodal learning across varied applications.
\end{itemize}

\section{Related Work}

\subsection{Multimodal Representation Learning}

Multimodal representation learning begins with building connections between vision and language modalities (bimodal representation learning)~\cite{radford2021learning, zhang2021cross,jia2021scaling,zhou2024few}. 
Particularly, CLIP~\cite{radford2021learning} learns deep semantic representations by matching vision concepts to linguistic inputs. This paradigm inspires a series of works to extend more modalities, \textit{e.g.}, audio-to-text~\cite{elizalde2023clap}, point-to-text~\cite{zhang2022pointclip}, and video-to-text~\cite{xu2021videoclip,luo2022clip4clip}.
These methods utilize pairwise contrastive learning~\cite{chen2020simple, he2020momentum, chen2020improved} to align two modality representations closer if they are from the same instance, while pushing away otherwise, thus building a joint embedding space. 
Building upon this bimodal paradigm, recent works introduce more modalities into a unified foundation model~\cite{chen2023valor,chen2023vast, cicchetti2024gramian, guo2023point, wang2024freebind, liu2024towards, liu2025continual, wang2024omnibind, girdhar2023imagebind, zhu2023languagebind, lyu2024unibind}.
Subtitles~\cite{luo2020univl, seo2021look, li2021value} and audio~\cite{guzhov2022audioclip, wu2022wav2clip, rouditchenko2020avlnet, ruan2023accommodating} are introduced and modeled together with vision and text. More training objectives, like next utterance prediction~\cite{seo2021look}, masked prediction~\cite{luo2020univl}, and modality pair matching~\cite{li2022blip, chen2023vast}, are adopted to further enhance the performance. Notably, VAST~\cite{chen2023vast} pioneers the omni-modality foundation model, involving vision, audio, subtitle, and text. 
Alongside these innovations, ImageBind~\cite{girdhar2023imagebind} builds upon CLIP and binds multiple modality representations, with vision modality being the anchor. 
By setting an anchor modality (\textit{e.g.}, language~\cite{zhu2023languagebind}, vision~\cite{girdhar2023imagebind}, and point cloud~\cite{guo2023point}), all the modalities will be aligned together through interactively contrastive learning. 
However, this privileging of one modality enforces a fixed representation to frame all modalities (\textit{i.e.}, single-modal centrism), inherently ignoring the complexity and richness of multimodalities.

Differently, GRAM~\cite{cicchetti2024gramian} proposes to align multimodal representations simultaneously, which spans a parallelotope, and minimizes its volume to achieve simultaneous alignment for all modalities. Nevertheless, its learning objective is implemented by contrastive learning, where the text modality serves as the anchor and is replaced to construct negative samples. A predefined anchor is still relied upon by GRAM. Additionally, the volume collapses when a singular value approaches zero, leading to unstable optimization.
Our method goes beyond this by strengthening the maximum singular value, aligning all modalities to the leading direction automatically, and theoretically revealing its potential to achieve full alignment.

\subsection{Principled Learning with SVD}

Singular value decomposition (SVD) is a fundamental matrix factorization technique~\cite{golub2013matrix, golub1971singular, wall2003singular, abdi2007singular,liu2025continual} with broad applications in machine learning~\cite{mathiasen2020if, levinson2020analysis,zhang2018stabilizing}. SVD has been extensively utilized in domains such as image processing~\cite{brand2002incremental, gong2000video, zhou2023star}, model compression~\cite{li2025adasvd,yang2020learning,wang2024svd,han2023svdiff}, and parameter initialization~\cite{meng2024pissa}. Despite its versatility, the potential of SVD in multimodal learning remains underexplored, presenting opportunities for novel applications and theoretical advancements.
Notably, contrastive loss minimization by gradient descent can be formulated as the SVD on a contrastive cross-covariance matrix, establishing the connection between SVD and multimodal contrastive learning~\cite{nakada2023understanding}. 
In addition to the theoretical analysis, recent work~\cite{kamboj2025leveraging} leverages SVD to construct the linear transformation from modality to representation, while being limited by a bimodal scenario. 
In this work, we further exploit the SVD analysis and establish a formal connection between singular values and full alignment in multimodal representation learning, which introduces a novel method that builds upon this theoretical insight.
\section{Preliminary}

\textbf{Bimodal alignment.}
Given multimodal data $\mathcal{X} = \{\mathbf{x}_i | i\in \mathbb{Z}^+, i< N\}$, $\mathbf{x}_i$ is a multimodal instance containing $k$ modalities, denoting $\mathbf{x}_i = \{\mathbf{x}_i^m | m \in \mathcal{M}\}$. Multimodal representation learning aims to learn the latent representation $\mathbf{z}_i^m \in \mathbb{R}^{d\times 1}$ from the corresponding multimodal data $\mathbf{x}_i^m$ through the encoder.
The latent representations $\{\mathbf{z}_i^m | m\in\mathcal{M}\}$ from the common instance $i$ are expected to be \textit{similar}, in other words, being retrievable from each other. 
Cross-modal retrieval offers insights where two modalities (\textit{e.g.}, $m_1$ and $m_2$) are aligned with pairwise contrastive learning, and the similarity is defined as the inner product between their representations:
\begin{align}
    \mathcal{L}^{m_1,m_2} = -\frac{1}{N}\sum_{i=1}^N\log\frac{\exp({\mathbf{z}_i^{m_1}}^\top \mathbf{z}_i^{m_2}/\tau)}{\sum_j^N\exp({\mathbf{z}_i^{m_1}}^\top \mathbf{z}_j^{m_2}/\tau)},
\end{align}
where $\tau$ is the temperature ratio and $N$ denotes the number of data pairs.
This bimodal alignment objective is also widely adopted for multimodal representation learning, including~\cite{zhu2023languagebind, chen2023vast, girdhar2023imagebind}.

\textbf{Multimodal alignment.}
The pairwise contrastive learning paradigm can also be extended to the multimodal scenario~($k>2$). For example, the training on $\{m_1, m_2, m_3\}$ can be decomposed to the training sequences of $\mathcal{L}^{m_1,m_2}$, $\mathcal{L}^{m_1,m_3}$, and $\mathcal{L}^{m_2,m_3}$~\cite{girdhar2023imagebind, chen2023vast}\footnote{Here we do not highlight the asymmetric property of $\mathcal{L}^{m_1, m_2}$. }. 
GRAM~\cite{cicchetti2024gramian} proposes to project all modalities to form a parallelotope with a small volume, which can be defined by the determinant of the Gram matrix, \textit{i.e.}, $\det(\mathbf{G}) = \det(\mathbf{Z}^\top \mathbf{Z})$. 
The performance improvement induced by aligning all modalities simultaneously motivates us to dive deeper into it.

\textbf{SVD and eigenvalues.}
Given an arbitrary matrix $\mathbf{X}\in \mathbb{R}^{n\times n'}$, we have $\mathbf{X} = \mathbf{U}\mathbf{\Sigma}\mathbf{V}^\top$ via SVD. Here $\mathbf{U}\in \mathbb{R}^{n'\times n'}$ and $\mathbf{V}\in\mathbb{R}^{n\times n}$ are unitary matrices, satisfying $\mathbf{U}\mathbf{U}^\top = \mathbf{1}$ and $\mathbf{V}\mathbf{V}^\top = \mathbf{1}$. Besides, $\mathbf{\Sigma}\in \mathbb{R}^{n'\times n}$ is the matrix with non-negative entries on the diagonal and zeros off the diagonal. The diagonal ones (singular values) can be represented as $ \sigma_1 \geq \sigma_2 \geq \dots \geq 0$, square roots of the eigenvalues of $ \mathbf{X}^\top \mathbf{X}$. The maximum eigenvalue $ \lambda_1 = \sigma_1^2 $ corresponds to the dominant singular direction.

\section{Principled Multimodal Representation Learning}

\subsection{Principled Learning}

\textbf{Alignment goal.} Given normalized representations $\{\mathbf{z}_i^m \mid m \in \mathcal{M}\}$ derived from a shared instance $i$, the alignment is typically quantified via pairwise inner products (\textit{e.g.}, cosine similarity)~\cite{radford2021learning, girdhar2023imagebind,zhu2023languagebind,chen2023vast}. Therefore, the objective is to maximize such similarity: $\arg\max a^{m_i,m_j}_{\bm{\theta}}:= (\mathbf{z}^{m_i})^\top\mathbf{z}^{m_j}, \forall \{m_i, m_j\} \subseteq \mathcal{M}$\footnote{We use superscript to denote modality indices and subscript for instance indices. Omitting the subscript indicates the same instance.}, where $\bm{\theta}$ denotes related parameters to be optimized.  We can also express this in a matrix form as: 
\begin{align}
    \operatorname*{argmax}_{\bm{\theta}} \|\mathbf{G}\|_F = \sqrt{\sum_{i,j} |a^{m_i,m_j}|^2}, &\nonumber
    \quad
    \mathbf{G}= \begin{bmatrix}
 1& a^{m_1,m_2} & \cdots & a^{m_1,m_k} \\
 a^{m_2,m_1} & 1 & \cdots & a^{m_2,m_k} \\
\vdots & \vdots & \ddots & \vdots \\
 a^{m_k,m_1} & a^{m_k,m_2} & \cdots & 1
\end{bmatrix},& 
\end{align}
where $\|\cdot\|_F$ is the Frobenius norm. The ideal case is that $\mathbf{z}^{m_1} = \mathbf{z}^{m_2}= \cdots= \mathbf{z}^{m_k}$, inducing maximum $\|\mathbf{G}\|_F$. Every entry in $\mathbf{G}$ equals 1. 
Notably, to avoid the extreme case where all the encoded representations are aligned with a common vector, there is typically an additional regularization, $a_{i,j} <  a_{i,i}$, across different instances. We leave this discussion in Section~\ref{sec:regularization}.

\begin{assumption}[]
 For a common instance, the alignment scores between pairs of modalities are nonnegative, \textit{i.e.}, $a^{m_i,m_2} \ge 0, \forall \{m_1,m_2\} \subseteq \mathcal{M}$. This implies that the angle formed by any paired multimodal representations is not obtuse if they are sourced from the same instance~\cite{cicchetti2024gramian}.
\end{assumption}

In the following, we first draw the connection between multimodal full alignment and the rank of the Gram matrix (\textit{cf.}, Lemma~\ref{lem:rank1}). Afterward, combined with the optimal rank-\textit{r} approximation (\textit{cf.}, Lemma~\ref{lem:rank_svd}), we derive our principled learning theory (\textit{cf.}, Theorem~\ref{thm:eigenvalue}) that motivates us to strengthen the maximum singular value to approach full alignment.

\textbf{\ding{172} Alignment and $\mathrm{rank}(\mathbf{G})$.} 
Considering the maximum $\|\mathbf{G}\|_F$, the ultimate goal, it satisfies that every element in $\mathbf{G}$ equals 1, and meanwhile $\mathrm{rank}(\mathbf{G}) = 1$. We have the following equivalence lemma. 

\begin{lem}[Full alignment $\Leftrightarrow$ Rank-1 Gram matrix]
\label{lem:rank1}
Let $\mathbf{G} \in \mathbb{R}^{k \times k}$ be a Gram matrix constructed from normalized modality representations $\{\mathbf{z}^m\}_{m=1}^k$, i.e., $\mathbf{G}_{i,j} = \langle \mathbf{z}^{m_i}, \mathbf{z}^{m_j} \rangle$ with $\|\mathbf{z}^{m_i}\| = 1$. Then the following are equivalent: (1) $\mathbf{G}_{i,j} = 1$ for all $i, j$, and (2) $\operatorname{rank}(\mathbf{G}) = 1$. See proof in Appendix~\ref{appendix:proof_rank1}.
\end{lem}

\vspace{-3mm}
\begin{tcolorbox}[colframe=gray!10, colback=gray!10, boxrule=0.2pt, arc=1pt, boxsep=0pt, left=2pt, right=2pt, top=2pt, bottom=2pt]
\textbf{Remark.} 
The proposed lemma establishes a fundamental connection between multimodal alignment and the rank of the Gram matrix. Therefore, we can transform the problem of multimodal alignment into achieving a rank-1 Gram matrix. A recent paper~\cite{cicchetti2024gramian} investigates the connection between the determinant of the Gram matrix and geometric interpretation. However, it fails to achieve full alignment because its objective can be satisfied with a collapsed dimension.
For a deeper understanding, we explore the connections between singular values and this work, and highlight the superiority of our method in Section~\ref{sec:further_analysis}. 
This equivalence offers a novel perspective that potentially inspires future research toward achieving full multimodal alignment.
\end{tcolorbox}
\vspace{2mm}

\textbf{\ding{173} Alignment and $\sigma_1$.} 
The goal is transformed to learn multimodal representations that yield the Gram matrix with rank 1. We propose a novel solution via SVD by maximizing the maximum singular value $\sigma_1$, supported by the following analysis (see Lemma~\ref{lem:rank_svd} and Theorem~\ref{thm:eigenvalue}).

\begin{lem}[Eckart-Young~\cite{eckart1936approximation}]
\label{lem:rank_svd}
The optimal rank-r approximation to $\mathbf{X}$, in a least-squares sense, is given by the rank-r SVD truncation $\tilde{\mathbf{X}}$:
\begin{align}
    \operatorname*{argmin}_{\tilde{\mathbf{X}},s.t.\operatorname{rank}(\tilde{\mathbf{X}})=r}\|\mathbf{X}-\tilde{\mathbf{X}}\|_{F}=\tilde{\mathbf{U}}\tilde{\mathbf{\Sigma}}\tilde{\mathbf{V}}^{\top}.
\end{align}
Here $\tilde{\mathbf{U}}$ and $\tilde{\mathbf{V}}$ denote the first $r$ leading columns of $\mathbf{U}$ and $\mathbf{V}$, and $\tilde{\mathbf{\Sigma}}$ contains the leading $r\times r$ sub-block of $\mathbf{\Sigma}$.
\end{lem}
\vspace{-3mm}

\begin{tcolorbox}[colframe=gray!10, colback=gray!10, boxrule=0.2pt, arc=1pt, boxsep=0pt, left=2pt, right=2pt, top=2pt, bottom=2pt]
\textbf{Remark.} 
Lemma~\ref{lem:rank_svd} reveals that the low-rank approximation can be optimally achieved via SVD, motivating a branch of work utilizing it for model compression~\cite{li2025adasvd,yang2020learning}. Despite the inspiration, in our context, the goal is to minimize $\|\mathbf{G} - \tilde{\mathbf{G}}\|_F$, where $\mathrm{rank}(\tilde{\mathbf{G}})=1$, by optimizing the learnable $\mathbf{Z}$. The optimal low-rank matrix ($\tilde{\mathbf{G}}$) is found, while $\mathbf{G} = \mathbf{Z}^\top \mathbf{Z}$ is under-resolved. 
\end{tcolorbox}

\begin{thm}[Principled learning]
\label{thm:eigenvalue}
Let $\mathbf{Z} = [\mathbf{z}^{m_1}, \dots, \mathbf{z}^{m_k}] \in \mathbb{R}^{d \times k}$ be a matrix of normalized modality representations from the same instance, i.e., $\|\mathbf{z}^{m_i}\| = 1$ for all $i$, and let $\sigma_1$ denote its maximum singular value. Then, we have (1) maximizing $\sigma_1$ maximizes the pairwise cosine similarities among $\{\mathbf{z}^m\}_{m=1}^k$, and (2) $\mathrm{rank}(\mathbf{G})=1$ is achieved if and only if $\sigma_1 = \sqrt{k}$.
See proof in Appendix~\ref{appendix:proof_eigenvalue}.
\end{thm}
\vspace{-3mm}

\begin{figure*}
    \centering
    \includegraphics[width=0.88\linewidth]{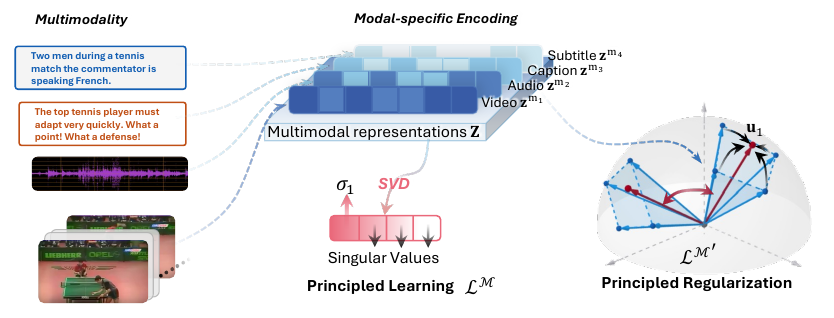}
    \caption{\textbf{The overall framework of PMRL.} Different modalities of the instance are encoded into multimodal representations $\mathbf{Z}$. PMRL utilizes SVD to obtain the maximum singular value $\sigma_1$ and maximizes it with the objective $\mathcal{L}^{\mathcal{M}}$. The leading directions (arrows in red) corresponding to $\sigma_1$ from different instances are regularized by $\mathcal{L}^{\mathcal{M}'}$.}
    \label{fig:framework}
\end{figure*}

\begin{tcolorbox}[colframe=gray!10, colback=gray!10, boxrule=0.2pt, arc=1pt, boxsep=0pt, left=2pt, right=2pt, top=2pt, bottom=2pt]
\textbf{Remark.} 
$\sigma_1$ reflects the strength of the leading direction of $\mathbf{u}_1$. By maximizing the $\sigma_1$, subject to the constraint $\sum_{i=1}^k \sigma_i^2 = k$, other singular values are minimized, finally aligning all representations $\mathbf{z}^m$ with the leading direction. 
Intuitively, this process adaptively identifies an optimal anchor for alignment at each training step, drawing all representations toward a common centroid. Figure~\ref{fig:framework} illustrates this concept for clarity. 
\end{tcolorbox}

\vspace{2mm}

\subsection{Singular Value Maximization for Multimodal Alignment}
Building upon the theoretical insights into multimodal alignment through Gram matrices and their spectral properties, we propose a novel learning objective that directly encourages alignment among heterogeneous modality representations. For each instance, we collect normalized embeddings from all available modalities and construct a compact representation matrix $\mathbf{Z}$. 
We apply SVD to extract the maximum singular value $\sigma_1$, which reflects the strength of the dominant alignment direction across modalities:
\begin{align}
     \text{SVD}(\mathbf{Z}) = \mathbf{U}\mathbf{\Sigma}\mathbf{V}, \quad
     \mathbf{\Sigma}  = \mathrm{diag}(\sigma_1, \sigma_2, \dots, \sigma_k). 
\end{align}
To enhance the prominence of the leading singular value during training, we introduce a softmax-based loss over the singular values:
\begin{align}
    \mathcal{L}^{\mathcal{M}} = -\frac{1}{N}\sum_{i=1}^{N}\log\frac{\exp[\sigma_1/\tau]}{\sum_{j}^{k}\exp[\sigma_j/\tau]},
    \label{eq:loss1}
\end{align}
where $\tau$ is a temperature parameter. 
This formulation treats the singular values as logits and encourages $\sigma_1$ to stand out relative to the rest, thereby promoting strong alignment.
The analysis on gradient propagation for improving alignment via singular value maximization is detailed in Appendix~\ref{appendix:gradient}. Below are the key insights that reveal its deeper significance.
(1) Unlike contrastive objectives, which optimize local similarity at the bimodal-level $(\mathbf{z}^{m_1})^\top \mathbf{z}^{m_2}$, this loss operates at the level of global covariance structure of $\mathbf{Z}$. PMRL goes beyond conventional contrastive learning~\cite{radford2021learning}, shifting MRL from isolated pairs to the collective behavior of modalities.
(2) Without predefined anchors, the model aligns modalities along a latent leading direction emerging from $\sigma_1$. 
By continuously amplifying the dominance through a differentiable competition among singular values, it fosters a self-discovering representation space.

\subsection{Instance-wise Regularization}
\label{sec:regularization}
To prevent degenerate solutions where all embeddings collapse to a single point or become misaligned across instances, we incorporate \textit{instance-wise contrastive regularization} that encourages separation between different instances:
\begin{align}
     \mathcal{L}^{\mathcal{M}'}  = -\frac{1}{N}\sum_{i=1}^{N}\log\frac{\exp [(\mathbf{u}_1^{(i)})^\top \mathbf{u}_1^{(i)}/\tau]}{\sum_{j}^{k}\exp [(\mathbf{u}_1^{(i)})^\top \mathbf{u}_1^{(j)}/\tau]},
     \label{eq:loss2}
\end{align}
where $\mathbf{u}_1$ corresponds to the maximum singular value $\sigma_1$, indicating the leading direction that all multimodal representations are aligned with. Furthermore, we employ the instance matching loss, which encourages the model to predict whether the multimodal data is matched or not as a binary question. The data obtained from all the encoders is concatenated and fed to a multimodal encoder. A two-layer multi-layer perceptron~(MLP) serves as the predictor that returns $\hat{y}$, the matching probability. We follow~\cite{chen2023vast, cicchetti2024gramian} with the hard negative mining strategy and employ the instance matching loss as follows: 
\begin{align}
    \mathcal{L}_{\mathrm{IM}} = \mathbb{E}_{(m_1,m_2,\dots,m_k)\sim\mathcal{M}}[y\log\hat{y} + (1-y)\log(1-\hat{y})].
    \label{eq:loss3}
\end{align}
The overall objective combines the alignment-driven singular value loss with auxiliary regularization terms:
\begin{align}
\mathcal{L} = \mathcal{L}^{\mathcal{M}} + \lambda_1\mathcal{L}^{\mathcal{M}'} + \lambda_2\mathcal{L}_{\text{IM}},
\label{eq:loss}
\end{align}
where $\lambda_1$ and $\lambda_2$ control the strength of regularizations. We set $\lambda_1=1$ and $\lambda_2=0.1$ by default. We provide the algorithm flow of our PMRL in Algorithm~\ref{alg:pmrl_training} to facilitate a comprehensive understanding and promote reproducibility.

\begin{algorithm*}[t]
\caption{PMRL: Principal Multimodal Representation Learning}
\label{alg:pmrl_training}
\begin{algorithmic}[1]
\REQUIRE 
    \textbf{Inputs} \\
    \quad Dataset $\mathcal{X} = \{(\mathbf{x}_i^{m_1}, \mathbf{x}_i^{m_2}, \dots, \mathbf{x}_i^{m_k})\}_{i=1}^N$ with $k = |\mathcal{M}|$ modalities per instance.\\
    \quad Encoder networks $\{f^m(\cdot; \bm{\theta}^m)\}_{m=1}^k$, parameterized by $\bm{\theta}^m$.\\
    \quad Temperature parameter $\tau > 0$, regularization weights $\lambda_1, \lambda_2$.
    
\ENSURE  Aligned multimodal representations.

\FOR{each training iteration}
    \STATE \textbf{Sample a batch of instances:} $\{\mathbf{x}_i^{m_1}, \mathbf{x}_i^{m_2}, \dots, \mathbf{x}_i^{m_k}\}_{i=1}^B$;
    
    \STATE $$
    \text{\ding{172} Modality-specific encoding:}\left\{\begin{aligned}
        &\text{Encode modality-specific embeddings: }
            \mathbf{z}^{m} = f^m(\mathbf{x}^{m}; \bm{\theta}^m), \quad \forall m \in \mathcal{M}; \\
        &\text{Normalize embeddings: }
            \mathbf{z}^{m} \leftarrow \frac{\mathbf{z}^{m}}{\|\mathbf{z}^{m}\|}, \quad \forall m \in \mathcal{M};\\
        &\text{Stack normalized embeddings into matrix: }
            \mathbf{Z} = [\mathbf{z}^{m_1}, \mathbf{z}^{m_2}, \dots, \mathbf{z}^{m_k}] \in \mathbb{R}^{d \times k};\quad\quad\quad \quad\quad
    \end{aligned}
    \right.
    $$

    \STATE \ding{173} Perform SVD on $\mathbf{Z}$: 
            $\text{SVD}(\mathbf{Z}) = \mathbf{U} \mathbf{\Sigma} \mathbf{V}^\top, \quad \mathbf{\Sigma} = \text{diag}(\sigma_{1}, \dots, \sigma_{k})$;
            
    \STATE \quad Extract leading singular value  $\sigma_{1}$ { and corresponding left singular vector } $\mathbf{u}_{1}$;
        
    $$
    \text{\ding{174} The core part of  PMRL:}\left\{\begin{aligned}
    &\text{Compute the alignment loss } \mathcal{L}^{\mathcal{M}}~\text{via softmax over singular values (see Eq.~(\ref{eq:loss1}))};\\
    &\text{Compute the contrastive regularization }\mathcal{L}^{\mathcal{M}'}~\text{using leading directions (see Eq.~(\ref{eq:loss2}))};\quad\quad\quad \quad\quad
    \end{aligned}
    \right.
    $$

    \STATE \ding{175} Generate matched/mismatched pairs for instance matching loss $\mathcal{L}_{\mathrm{IM}}$ (see Eq.~(\ref{eq:loss3}));

    \STATE \ding{176} Combine losses with weighting coefficients:
    $
        \mathcal{L} = \mathcal{L}^{\mathcal{M}} + \lambda_1 \mathcal{L}^{\mathcal{M}'} + \lambda_2 \mathcal{L}_{\mathrm{IM}};
    $

    \STATE \ding{177} Update parameters $\bm{\theta}$ via gradient descent:
    $
        \bm{\theta} \leftarrow \bm{\theta} - \eta \nabla_{\bm{\theta}} \mathcal{L};
    $
\ENDFOR

\STATE \textbf{Return:} Optimized encoder parameters $\bm{\theta}$ that encourage fully aligned multimodal representations.
\end{algorithmic}
\end{algorithm*}

\subsection{Further Analysis}
\label{sec:further_analysis}

\textbf{Connecting to the volume of the Gram matrix.} 
Prior work~\cite{cicchetti2024gramian} investigates minimizing the determinant of the Gram matrix, which can be interpreted geometrically, \textit{i.e.}, the volume of the $k$-dimensional parallelotope formed by multimodal representations. Unfortunately, the theoretical connection between the volume and multimodal alignment is still unexplored. Here we highlight insights shared with GRAM~\cite{cicchetti2024gramian} through singular value analysis while distinguishing our work through approaching full alignment. Specifically, the volume proposed by GRAM can be represented by the product of singular values as well: 
\begin{align}
\text{Vol}(\mathbf{G}) = \sqrt{\det \mathbf{G}} = \sqrt{\det \mathbf{Z}^\top\mathbf{Z}} = \prod_{i=1}^k \sigma_i.
\end{align}
Afterward, the volume achieves its minimum value of zero if and only if at least one of the singular values $\{\sigma_i\}_{i=1}^k$ is zero. In other words, by optimizing the minimum singular value $\sigma_k$ to zero, the volume of the $k$-dimensional parallelotope reaches zero as well. In this case, the Gram matrix rank can remain larger than 1, preventing full alignment (see Lemma~\ref{lem:rank1}). We also illustrate the trend of singular values during training in Section~\ref{sec:further_empirical_analysis}. Geometrically, this corresponds to the parallelotope \textit{collapsing} to $k-1$ dimensions. The collapsed volume is no longer optimized. In practice, GRAM is also achieved with a pre-defined anchor for contrastive learning. In comparison, we encourage the multimodal alignment in an \textit{anchor-free} manner.

\textbf{Robustness to noise.} PMRL showcases robustness against noise in input data and labels, as supported by prior works~\cite{kim2023rank,dusenberry2020efficient}. 
Noise, \textit{e.g.}, Gaussian perturbations, disrupts the generation of multimodal representations, complicating alignment estimation. SVD effectively filters noisy data~\cite{gavish2017optimal, epps2019singular} and recovers rank-1 matrices~\cite{kim2023rank}. Additionally, noisy labels can destabilize learning processes~\cite{xia2021sample,xia2020part}, but SVD-extracted low-rank matrices alleviate these negative effects~\cite{dusenberry2020efficient, kodge2025sap}. These findings highlight PMRL’s robustness, approaching rank-1 matrices to promote full alignment.

\section{Experiments}

\begin{table*}[t]
\centering
\caption{Multimodal text-to-video (T$\rightarrow$V) and video-to-text (V$\rightarrow$T) retrieval results~(\%) in the zero-shot setting, in terms of Recall@1 (R@1). Increment points are computed compared with VAST. }
\label{tab:vt_zero}
\begin{adjustbox}{width=\textwidth,center}
\begin{tabular}{@{}l|ll|ll|ll|ll@{}}
\toprule
 & \multicolumn{2}{c|}{MSR-VTT} & \multicolumn{2}{c|}{DiDeMo} & \multicolumn{2}{c|}{ActivityNet}  & \multicolumn{2}{c}{VATEX}   \\  & T$\rightarrow$V    & V$\rightarrow$T   & T$\rightarrow$V     & V$\rightarrow$T  & T$\rightarrow$V    & V$\rightarrow$T  & T$\rightarrow$V    & V$\rightarrow$T    \\ \midrule 
Fronzen \cite{bain2021frozen}                    & 18.7   &  -  & 21.1     &  - & -   & -  & -&-\\
UMT \cite{liu2022umt}                         & 33.3   &  -  & 34.0     &  - & 31.9   & -  & -&-\\
UMT-L \cite{li2023unmasked}  & 40.7  & 37.1     & 48.6      & 49.9     & 41.9  & 39.4 & -& -  \\
OmniVL \cite{wang2022omnivl} & 42.0    &  -  & 40.6    & -  & -   & -  & -&-\\
TVTSv2 \cite{zeng2023tvtsv2}  & 38.2       &    -   & 34.6      &  -    & -     & -  & -&-\\
ViCLIP \cite{wang2023internvid}  & 42.4  & 41.3      & 18.4      & 27.9     & 15.1    & 24.0 & - & - \\
VideoCoCa \cite{yan2022videococa}   & 34.3    & 64.7   & -         & -     & 34.5     & 33.0 & 53.2 & 73.6 \\
Norton \cite{lin2023multi}                    &   10.7  &    &     -     &    -  &    -  &  - & - & - \\ 
ImageBind \cite{girdhar2023imagebind}  & 36.8    &  -  & -        & -     & -    & -  & -&-\\
InternVideo-L \cite{wang2022internvideo}  & 40.7 &    39.6   & 31.5     & 33.5     & 30.7      & 31.4 & 49.5 & 69.5 \\
HiTeA \cite{ye2023hitea}  & 34.4 &    -   & 43.2     & -     & -      & -  & -&-\\
mPLUG-2 \cite{xu2023mplug}  & 47.1 &    -   & 45.7     & -     & -      & -  & -&-\\
VideoPrism-b \cite{zhao2024videoprism}  & 51.4     & 50.2   & -    & -  & 49.6      &  47.9 & 62.5 & 77.1\\
LanguageBind \cite{zhu2023languagebind}  & 44.8     & 40.9   & 39.9    & 39.8  & 41.0      &  39.1 & - & - \\ \midrule 
VAST \cite{chen2023vast}   & 50.5   & 48.8   & 46.4 & 45.3 & 51.7 & 48.8& 75.9 & 74.8 \\
GRAM \cite{cicchetti2024gramian} & 51.5 \textcolor{lightblue}{(+1.0)}    & 51.5 \textcolor{lightblue}{(+1.0)}   & 49.8 \textcolor{lightblue}{(+2.6)}   & 48.5 \textcolor{lightblue}{(+3.2)}  & 54.5 \textcolor{lightblue}{(+2.8)}  & 48.3 \textcolor{lightblue}{(-0.5)}  & 77.5 \textcolor{lightblue}{(+1.6)}  & 74.7 \textcolor{lightblue}{(-0.1)} \\

\midrule
\textbf{PMRL (Ours)}  & \textbf{54.5 \textcolor{darkblue}{(+4.0)}}   & \textbf{52.4 \textcolor{darkblue}{(+3.6)}}   & \textbf{50.6 \textcolor{darkblue}{(+4.2)}}     &  \textbf{48.4 \textcolor{darkblue}{(+3.1)}} & \textbf{56.0 \textcolor{darkblue}{(+5.3)}}    & \textbf{49.6 \textcolor{darkblue}{(+0.8)}} & \textbf{80.5 \textcolor{darkblue}{(+4.6)}} & \textbf{75.2 \textcolor{darkblue}{(+0.4)}}\\
 \bottomrule
\end{tabular}
\end{adjustbox}
\vspace{-1mm}
\end{table*}

\subsection{Experimental Setups}

\textbf{Datasets.}
VAST-150K~\cite{cicchetti2024gramian}, a downsized version of VAST-27M~\cite{chen2023vast}, is utilized for the multimodal training. This dataset involves four modalities, including vision, audio, subtitle, and text (\textit{i.e.}, caption). 
For the downstream evaluation, we utilize MSR-VTT~\cite{chen2011collecting}, DiDeMo~\cite{anne2017localizing}, ActivityNet~\cite{krishna2017dense}, and VATEX~\cite{wang2019vatex} for text-video retrieval, and AudioCaps~\cite{kim2019audiocaps} and Clotho~\cite{drossos2020clotho} for text-audio retrieval. In addition, to demonstrate the broader potential of our method, we incorporate the ABIDE (Autism Brain Imaging Data Exchange)~\cite{craddock2013neuro} dataset, a brain imaging dataset for autism classification, covering three modalities (\textit{i.e.}, fMRI, sMRI, and text).
PMRL is built upon VAST and employs a continual pre-training strategy to evaluate its effectiveness, following~\cite{cicchetti2024gramian}. Therefore, we utilize VAST-150K to re-boost its zero-shot capabilities, and split downstream datasets for fine-tuning PMRL for specific tasks. All the downstream datasets involve over two modalities.
See more details of the datasets in Appendix~\ref{appendix:datasets}. 

\textbf{Baselines and evaluation metrics.}
We select extensive baselines in our comparison, including 
Frozen \cite{bain2021frozen}, 
UMT \cite{liu2022umt}, 
UMT-L \cite{li2023unmasked}, 
OmniVL \cite{wang2022omnivl}, 
TVTSv2 \cite{zeng2023tvtsv2}, 
CLIP4Clip \cite{luo2022clip4clip},
ViCLIP \cite{wang2023internvid}, 
VideoCoCa \cite{yan2022videococa}, 
Norton \cite{lin2023multi}, 
ImageBind \cite{girdhar2023imagebind}, 
InternVideo-L \cite{wang2022internvideo},
HiTeA \cite{ye2023hitea}, 
mPLUG-2 \cite{xu2023mplug}, 
VALOR-L \cite{liu2024valor}, 
TEFAL \cite{ibrahimi2023audio}, 
Bimodal T2M \cite{arora2024text}, 
T-MASS \cite{wang2024text}, 
vid-TLDR \cite{choi2024vid},
VideoPrism-b \cite{zhao2024videoprism}, 
LanguageBind \cite{zhu2023languagebind}, 
AVFIC~\cite{nagrani2022learning},
VIP-ANT~\cite{zhao2021connecting},
VAST \cite{chen2023vast}, 
and 
GRAM \cite{cicchetti2024gramian} (more details are shown in Appendix~\ref{appendix:baselines}). Wherein, GRAM serves as the state-of-the-art (SOTA) method. 
In this main comparison, we conduct the evaluation in the zero-shot setting. We also implement the fine-tuning setting in multimodal text-to-video retrieval, following~\cite{cicchetti2024gramian}. 
We utilize \textit{Recall} as the retrieval metric. Note that we implement VAST's evaluation algorithm, which uses a conventional cosine similarity-based method.
For ABIDE, we use the training datasets to align new modalities (\textit{e.g.}, fMRI and sMRI). Here we select 
AE-FCN~\cite{rakic2020improving},
GCN~\cite{parisot2018disease},
VanillaTF,
BrainNetCNN~\cite{kawahara2017brainnetcnn},  and 
BrainNetTF~\cite{kan2022brain} as baselines for classification performance comparison. VAST~\cite{chen2023vast} and GRAM~\cite{cicchetti2024gramian} are also specialized for this task to ensure a fair comparison with our PMRL. \textit{AUC} and \textit{Accuracy} serve as the classification metric. The baselines and relevant metrics are detailed in Appendices~\ref{appendix:baselines} and~\ref{appendix:metrics}. 

\vspace{4.2mm}
\textbf{Model architecture and hyperparameters.} The PMRL model is built upon VAST~\cite{chen2023vast} with the same architecture in our main comparison. Specifically, the vision, audio, and text encoders are implemented via EVAClip-ViT-G~\cite{sun2023eva}, BEATs~\cite{chen2023beats}, and BERT-B~\cite{devlin-etal-2019-bert}, respectively. We continue optimizing the parameters of VAST with our proposed objective.  For autism evaluation, we implement our PMRL model (also VAST and GRAM) with BrainNetTF~\cite{kan2022brain} for the fMRI modality, a multi-layer perceptron~(MLP) module for the sMRI modality, and BERT for the textual modality. Built on generated representations, we add an MLP classifier. 
{
To extend our method to additional modalities and diverse tasks, we implement PMRL on top of ImageBind, introducing additional parameters after the feature encoder. We trained this version on VAST-150K for zero-shot evaluation. For hyperspectral imaging tasks, we implement PMRL using a convolutional neural network with mean pooling to process LiDAR and HSI data. And text labels are directly represented as learnable embeddings. To ensure a fair comparison, we implement VAST and GRAM using the same architecture as PMRL.}
By default, we set the learning rate to 2$\times10^{-5}$, the batch size to 64, and train the model for one epoch. We utilize AdamW~\cite{loshchilov2017decoupled} as the optimizer and a linear warmup scheduler. Experiments are conducted in a device equipped with 4$\times$NVIDIA H100-80GB GPUs. Detailed hyperparameter settings and model architectures can be found in Appendix~\ref{appendix:hyperparameters} and Appendix~\ref{appendix:model_architecure}, respectively. We also provide the pseudocode to facilitate reproducibility, as shown in Appendix~\ref{appendix:pseudocode}.

\subsection{Main Results}

\begin{table*}[t]
\centering
\caption{Multimodal text-to-video (T$\rightarrow$V) and video-to-text (V$\rightarrow$T) retrieval results (\%) in the finetuning setting, in terms of Recall@1 (R@1). Increment points are computed compared with VAST\protect\footnotemark. }
\label{tab:vt_finetune}
\begin{adjustbox}{width=\textwidth,center}
\begin{tabular}{@{}l|ll|ll|ll|ll@{}}
\toprule
  & \multicolumn{2}{c|}{MSR-VTT} & \multicolumn{2}{c|}{DiDeMo} & \multicolumn{2}{c|}{ActivityNet}  & \multicolumn{2}{c}{VATEX} \\ 
 & T$\rightarrow$V    & V$\rightarrow$T   & T$\rightarrow$V     & V$\rightarrow$T  & T$\rightarrow$V    & V$\rightarrow$T  & T$\rightarrow$V    & V$\rightarrow$T   \\ \midrule
UMT-L \cite{li2023unmasked} & 58.8$^*$  & 58.6$^*$     & 70.4$^*$ & 65.7$^*$     & 66.8$^*$  & 64.4$^*$  & 72.0$^*$ & 86.0$^*$ \\
CLIP4Clip \cite{luo2022clip4clip} & 44.5  & 45.9     & 43.4 & 43.6     & 40.5  & 41.6  & 55.9 & 78.3 \\
ViCLIP \cite{wang2023internvid}  & 52.5  & 51.8      & 49.4      & 50.2     & 49.8    & 48.1 & - & - \\
InternVideo-L \cite{wang2022internvideo}  & 55.2$^*$ &    57.9$^*$   & 57.9$^*$     & 59.1$^*$     & 62.2$^*$      & 62.8$^*$ & 71.1$^*$ & 87.2$^*$ \\
HiTeA \cite{ye2023hitea}  & 46.8 &    -   & 56.5     & -     & -      & -  & -&-\\
mPLUG-2 \cite{xu2023mplug}  & 53.1 &    -   & 56.4     & -     & -      & -  & -&-\\
VALOR-L \cite{liu2024valor}  & 54.4 &    -   & 57.6     & -     & 63.4      & -  & 76.9&-\\
TEFAL \cite{ibrahimi2023audio}  & 52.0 &    -   & -     & -     & -      & -  & 61.0 &-\\
Bimodal T2M \cite{arora2024text}  & 36.8 &    -   & -     & -     & -      & -  & - &-\\
T-MASS \cite{wang2024text}  & 52.7 &    -   & 53.3     & -     & -      & -  & 65.6 &-\\
vid-TLDR \cite{choi2024vid}  & 58.5$^*$   &  -  & 70.4$^*$ & - & 65.2$^*$ & -  & - & -  \\
\midrule 
VAST \cite{chen2023vast}  & 64.4   &  64.3  & 68.4   & 65.4 & 68.1   &  65.4  & 83.1   &  81.3  \\

GRAM \cite{cicchetti2024gramian} & 60.0 \textcolor{lightblue}{(-4.4)}    & 61.8 \textcolor{lightblue}{(-2.5)}   & 68.7 \textcolor{lightblue}{(+0.3)}   & 65.7 \textcolor{lightblue}{(+0.3)}  & 67.6 \textcolor{lightblue}{(-0.5)}  & 65.0 \textcolor{lightblue}{(-0.4)}  & 82.5 \textcolor{lightblue}{(-0.6)}  & 80.6 \textcolor{lightblue}{(-0.7)} \\

\midrule

\textbf{PMRL (Ours)}  & \textbf{61.2 \textcolor{darkblue}{(-3.2)}}   & \textbf{60.7 \textcolor{darkblue}{(-3.6)}}   & \textbf{70.2 \textcolor{darkblue}{(+1.8)}}     &  \textbf{66.4 \textcolor{darkblue}{(+1.0)}} & \textbf{68.2 \textcolor{darkblue}{(+0.1)}}    & \textbf{66.4 \textcolor{darkblue}{(+1.0)}} & \textbf{84.1 \textcolor{darkblue}{(+1.0)}} & \textbf{83.4 \textcolor{darkblue}{(+1.1)}}\\
 \bottomrule
\end{tabular}
\end{adjustbox}
\end{table*}

\begin{table*}[t]
    \centering
    \begin{minipage}[t]{0.6\linewidth}
        \centering
        \caption{Multimodal text-to-audio retrieval results (\%) in the zero-shot setting, in terms of Recall@1 (R@1) and 10 (R@10) scores. 
        }
        \label{tab:at}
        \begin{adjustbox}{width=\textwidth, center}
        \begin{tabular}{l|ll|ll}
        \toprule
        &  \multicolumn{2}{c|}{AudioCaps} & \multicolumn{2}{c}{Clotho} \\
        & R@1  & R@10   & R@1 & R@10         \\ \midrule
        AVFIC \cite{nagrani2022learning}                      & 8.7       & 37.7   & 3.0 & 17.5            \\
        AVFIC \cite{nagrani2022learning}                      & 10.6       & 45.2   & - & -            \\
        VIP-ANT  \cite{zhao2021connecting} & 27.7      & 37.7     & - & -        \\
        ImageBind  \cite{girdhar2023imagebind}& 9.3       & 42.3         & 6.0 & 28.4      \\
        LanguageBind \cite{zhu2023languagebind}& 19.7       & 67.6   & 16.7  & 52.0           \\\midrule 
        VAST  \cite{chen2023vast}     & 33.7      & 77.1  & 12.4 & 36.4 \\
        GRAM \cite{cicchetti2024gramian}        &  {34.6} \textcolor{lightblue}{(+0.9)}  & {77.4} \textcolor{lightblue}{(+0.3)}  & {15.9} \textcolor{lightblue}{(+3.5)}  & {43.6} \textcolor{lightblue}{(+7.2)}  \\
         \midrule
        \textbf{PMRL (Ours)}        &  \textbf{36.1 \textcolor{darkblue}{(+2.4)}}  & \textbf{75.9 \textcolor{darkblue}{(-1.2)}}  & \textbf{16.8 \textcolor{darkblue}{(+4.4)}}  & \textbf{44.0 \textcolor{darkblue}{(+7.6)}}   \\\bottomrule
        \end{tabular}
\end{adjustbox}
    \end{minipage}
    \hfill
    \begin{minipage}[t]{0.382\linewidth}
        \centering
        \caption{Multimodal autism classification results  (\%) in terms of AUC and ACC.
        }
        \label{tab:abide}
        \begin{adjustbox}{width=\textwidth,center}
        \begin{tabular}{l|ll}
        \toprule
        &  \multicolumn{2}{c}{ABIDE} \\ 
                                    & AUC         & ACC      \\ \midrule 
        AE-FCN \cite{rakic2020improving}                      & 78.9      & 69.4  \\
        GCN \cite{parisot2018disease}                      & 60.0     & 56.8     \\
        VanillaTF & 76.1      & 68.2       \\
        BrainNetCNN  \cite{kawahara2017brainnetcnn}& 73.6       & 67.9        \\
        BrainNetTF \cite{kan2022brain}& 78.7       & 70.6        \\ \midrule 
        VAST \cite{chen2023vast}      &  {79.2}  & 71.8  \\
        GRAM  \cite{cicchetti2024gramian}     &  {63.9}   & {60.6} \\
         \midrule
        \textbf{PMRL (Ours)}        &  \textbf{80.5 \textcolor{darkblue}{(+1.8)}} & \textbf{73.2 \textcolor{darkblue}{(+1.4)}}  \\\bottomrule
        \end{tabular}
        \end{adjustbox}
    \end{minipage}
\end{table*}

In this subsection, we mainly explore the performance of PMRL via developed evaluations, like, cross-modal retrieval and vision-audio classification, comparing it against existing baselines. We further demonstrate its broader impact on scientific applications, exemplified by autism diagnosis and hyperspectral imaging classification tasks. Finally, we validate PMRL’s capability to handle diverse modalities through evaluations on depth and tactile datasets.

\vspace{0.2mm}
\textbf{Multimodal cross-modal retrieval.} We evaluate the retrieval performance to indicate the alignments between modalities. It is well-established for several MRL methods, and typically focuses on text-video retrieval (\textit{i.e.}, 
\footnotetext{$^*$\small Finetuning and evaluation with 12 frames.}
Table~\ref{tab:vt_zero} for the zero-shot setting, while Table~\ref{tab:vt_finetune} for the fine-tuning setting), and text-to-audio retrieval (see Table~\ref{tab:at}). We follow the conventional cosine-based similarity metric for retrieval evaluation~\cite{chen2023vast}\footnote{We adapt the baseline GRAM for cosine-based evaluation to ensure a fair comparison.}. From these results, we have the following observations.
For text-video retrieval on four datasets, the PMRL model achieves substantial performance improvements, outperforming VAST by up to 5.3\% in retrieval metrics. Furthermore, the results also showcase that PMRL surpasses GRAM in both settings. More results are detailed in Appendix~\ref{appendix:tv_full}.
For multimodal text-to-audio retrieval across two datasets, as shown in Table~\ref{tab:at}, PMRL brings up to 7.6\% performance boost to VAST, and outperforms GRAM as well. 
Overall, the enhancement for the maximum singular value, the core objective of PMRL, brings performance boosts. The improvements indicate that a better multimodal representation can be learned from our proposed method. We can attribute it to our principled learning, exploring the fundamental goal of multimodal alignment and approaching it via resolving the algebraic problem. 
Specifically, we observe that GRAM performs worse compared to VAST in some cases. Its learning objective is specialized for volume as a measure of multimodal alignment, which is incompatible with the widely adopted cosine similarity. Despite the improvements brought by PMRL for most cases, we find that both GRAM and PMRL perform worse than VAST if we directly fine-tune the model on the MSR-VTT dataset. One possible reason is that MSR-VTT is cleaner and exhibits more manifest correlations between video and text modalities, which can be easily captured by vision-text specialized methods, like VAST. For datasets curated from wild (\textit{e.g.}, DiDeMo), we can observe a relatively better performance of PMRL.  
We also provide more analysis, like ablation studies and any modality retrieval results in Section~\ref{sec:further_empirical_analysis}, which offers more insights about PMRL. 

\textbf{Multimodal autism classification.} 
We demonstrate the broader impact of PMRL, especially focusing on multimodal autism classification. Table~\ref{tab:abide} provides the evaluation results on ABIDE concerning AUC and ACC metrics. We adopt multimodal representation learning objectives for the autism classification. Therefore, we introduce VAST, GRAM, and PMRL  with a classification loss. Compared to previous methods, \textit{e.g.}, BrainNetTF, more modalities benefit the performance improvements. Among these multimodal methods, PMRL outperforms others on both metrics (\textit{e.g.,} 3.6\% and 1.9\% improvements \textit{v.s.} VAST). Despite using modalities like fMRI and sMRI, PMRL shows its strong potential to enhance performance in more applications. We also observe a particularly low performance of GRAM when we conduct the training from scratch. Bolstered by the analysis on GRAM's volume collapse in Section~\ref{sec:further_analysis}, we can attribute it to its optimization leading to an incorrect direction to align multimodal representation, especially for the model with random initialization. We provide the singular value trends for both GRAM and PMRL in the next section (\textit{cf.}, eigenvalues analysis) to illustrate PMRL's more stable and goal-oriented learning procedure. 

{\textbf{Vision-audio classification.} 
We demonstrate performance improvements over baselines in zero-shot vision and audio classification. To conduct this evaluation, we utilize cross-modal retrieval between text labels and other modalities without training an additional classifier. The results are shown in Table~\ref{tab:0-shot_classification}. Compared to ImageBind, we observe consistent improvements from VAST, GRAM, and PMRL. Notably, PMRL significantly outperforms the other methods.}

{\textbf{Multimodal hyperspectral imaging classification.} 
In addition to the medical adaptation of PMRL, we evaluate its performance in the hyperspectral imaging domain. Specifically, we also employ cross-modal retrieval instead of an additional classifier head on the Houston13 dataset, which involves three modalities: LiDAR, HSI, and text. We average the LiDAR and HSI features for classification and report the results in Table~\ref{tab:houston}. We utilize three metrics for evaluation, involving Overall Accuracy (OA), Average Accuracy (AA), and Kappa ($\kappa$). PMRL demonstrates superior adaptability to broader domains, outperforming baselines by a significant margin. }

\begin{wraptable}{r}{0.6\textwidth}
\centering
\vspace{-3mm}
\caption{{Performance comparison for few-shot retrieval across different models.}}
\label{tab:few-shot_classification}
\begin{adjustbox}{width=\linewidth,center}
\begin{tabular}{l!{\vrule width \lightrulewidth}cc!{\vrule width \lightrulewidth}cc!{\vrule width \lightrulewidth}cc!{\vrule width \lightrulewidth}cc} 
\toprule
\multirow{3}{*}{Method} & \multicolumn{4}{c!{\vrule width \lightrulewidth}}{NYUDv2}                                                       & \multicolumn{4}{c}{TVL}                                                                    \\
                        & \multicolumn{2}{c!{\vrule width \lightrulewidth}}{V$\to$D} & \multicolumn{2}{c!{\vrule width \lightrulewidth}}{D$\to$V} & \multicolumn{2}{c!{\vrule width \lightrulewidth}}{V$\to$TA} & \multicolumn{2}{c}{TA$\to$V}         \\
                        & R@1           & R@10                                   & R@1           & R@10                                   & R@1            & R@10                                   & R@1            & R@10            \\ 
\midrule
ImageBind               & 0.46          & 4.13                                   & 0.00             & 3.67                                   & 0.25           & 4.23                                   & 0.00              & 2.49            \\
VAST                      & 3.21          & 21.41                                  & 3.21          & 23.24                                  & 10.2           & 34.83                                  & 1.74           & 13.18           \\
GRAM                    & 0.00             & 2.14                                   & 0.15          & 1.99                                   &  0.50  &            5.22 &           0.50  &            3.48       \\ 
\midrule
\textbf{PMRL (ours)}    & \textbf{5.05} & \textbf{25.23}                         & \textbf{5.96} & \textbf{29.66}                         & \textbf{17.66} & \textbf{39.05}                         & \textbf{21.89} & \textbf{41.04}  \\
\bottomrule
\end{tabular}
\end{adjustbox}
\end{wraptable}
\textbf{Adapt more modalities.} 
PMRL is capable of handling diverse modalities beyond vision, audio, and text. We evaluate PMRL on the novel modalities of depth (NYUDv2) and touch (TVL), with few-shot retrieval results reported in Table~\ref{tab:few-shot_classification}. ImageBind generally struggles to process these novel modalities, and fine-tuning leads to only marginal improvements or even performance drops for GRAM. In contrast, VAST and PMRL demonstrate significant gains. Notably, PMRL surpasses VAST by a remarkable margin.

\begin{table*}[t]
    \centering
    \begin{minipage}[t]{0.53\linewidth}
    \caption{{Performance comparison for zero-shot classification across different models.}}
    \label{tab:0-shot_classification}
    \centering
    \begin{tabular}{l|ccc|c}
        \toprule
         Dataset            &   ImageBind &    VAST &   GRAM &   \textbf{PMRL (ours)} \\
        \midrule
         VGGSound  &       30.98 & 33.58 &  \underline{34.58} &  \textbf{36.43} \\
         UCF101    &        69.31 & \underline{73.47} &  71.78 &  \textbf{74.06} \\
         ImageNet  &        69.66 & 69.70  &  \underline{70.71} &  \textbf{72.10}  \\
        \bottomrule
        \end{tabular}
    \end{minipage}
    \hfill
    \begin{minipage}[t]{0.42\linewidth}
        \caption{{Performance comparison on Houston13 of VAST, GRAM, and PMRL methods.}}
      \label{tab:houston}
      \centering
      \begin{tabular}{l|ccc}
        \toprule
         & OA & AA  & $\kappa$  \\
        \midrule
        VAST & 7.17 & 9.04 & 2.74 \\
        GRAM & 13.80 & 12.36 & 8.08 \\
        \textbf{PMRL (ours)} & \textbf{26.51} & \textbf{28.65} & \textbf{20.49} \\
        \bottomrule
      \end{tabular}
  \end{minipage}
\end{table*}

\begin{figure}
    \centering
    \includegraphics[width=0.98\linewidth]{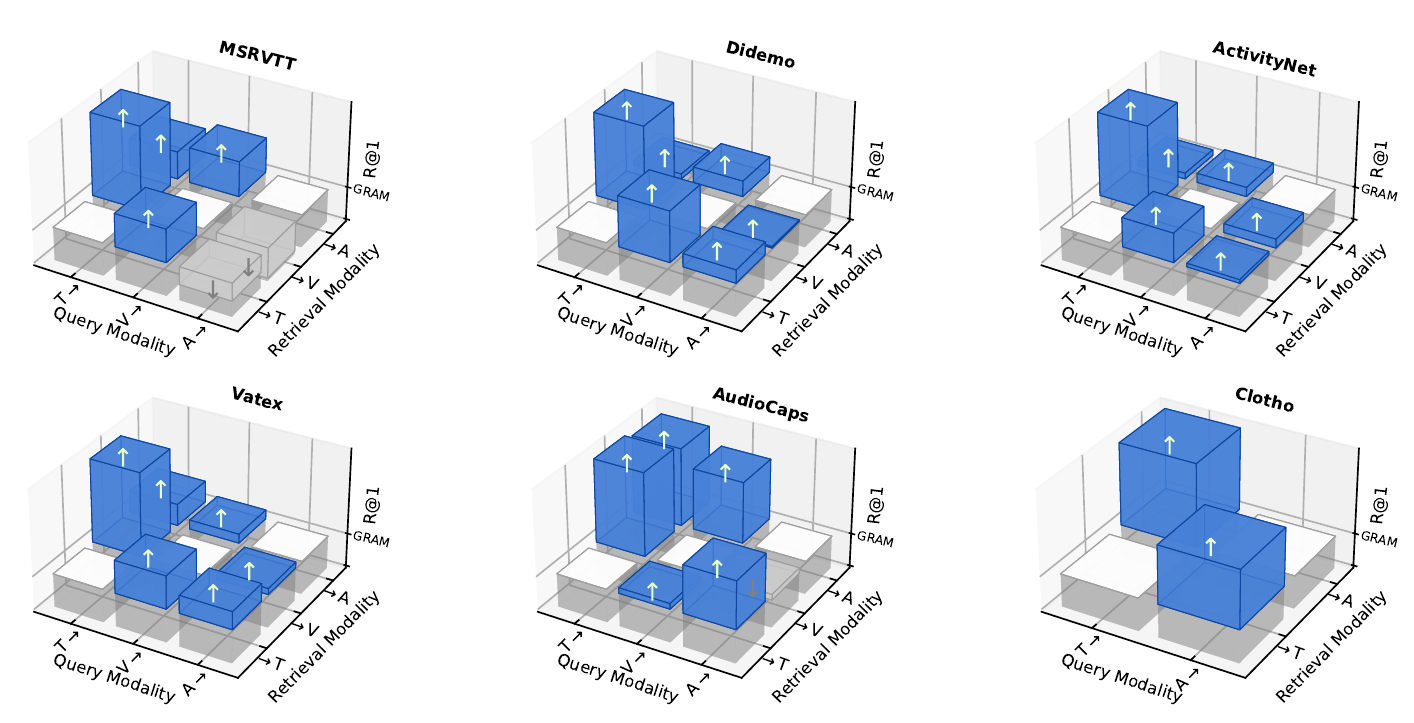}
    \caption{\textbf{Performance comparison for any modality retrieval across 6 benchmark datasets.} PMRL is compared with GRAM in terms of Recall@1. Blue regions highlight where PMRL outperforms GRAM, while gray regions indicate the opposite. Diagonal regions (colored in white) represent self-modal retrieval, which is not meaningful.}
    \label{fig:anchor-free}
    \vspace{0.8mm}
\end{figure}

\subsection{Further Empirical Analysis}
\label{sec:further_empirical_analysis}

To elucidate PMRL, we perform a comprehensive analysis supported by further empirical results. We conduct an ablation study in PMRL's design and evaluate retrieval performance across any modalities. We also track changes in singular values during training and examine the efficacy of instance-wise regularization. 
{We illustrate the distribution of $\sigma_1$ to confirm our theoretical assumption regarding the rank-1 Gram matrix. To validate the practical application of PMRL, we analyze its efficiency in terms of time and memory costs, with a specific focus on SVD computation.}
We interpret modality contributions to alignment using eigenvectors. 
{To validate the rationale behind our learning objective, we compare PMRL against a designed variant that encourages a higher rank, which indicates lower alignment.}
Finally, we evaluate PMRL's robustness to noise.

\begin{wrapfigure}{r}{0.6\textwidth}
    \centering
    \vspace{-6mm}
    \includegraphics[width=\linewidth, clip, trim=0pt 0pt 0pt 0pt]{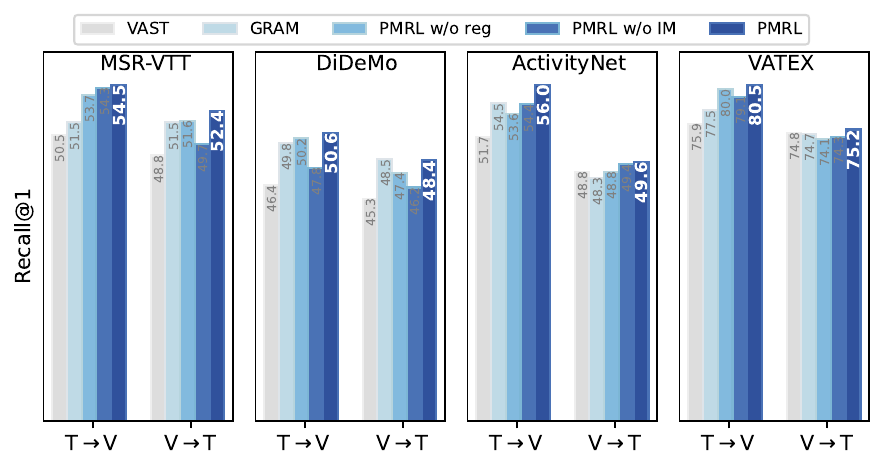}
    \caption{\textbf{The ablation study across 4 datasets in terms of Recall@1.} The instance-wise regularization loss (PMRL w/o reg) and instance matching loss (PMRL w/o IM) are canceled from PMRL and then compared with VAST and GRAM. }
    \label{fig:ablation}
    \vspace{-10mm}
\end{wrapfigure}
\textbf{Ablation study.}
To evaluate the efficacy of our proposed PMRL, we conduct the ablation study on our core designs, including principled learning on singular values ($\mathcal{L}^{\mathcal{M}}$) and principled regularization ($\mathcal{L}^{\mathcal{M}'}$). 
We report the results on four datasets, as shown in Figure~\ref{fig:ablation}. Without regularization (\textit{i.e.}, w/o reg or w/o IM), PMRL's performance declines across all scenarios. The integration of the proposed objectives yields great synergy to enhance multimodal representations.

\begin{figure*}
    \centering
    \includegraphics[width=0.98\linewidth]{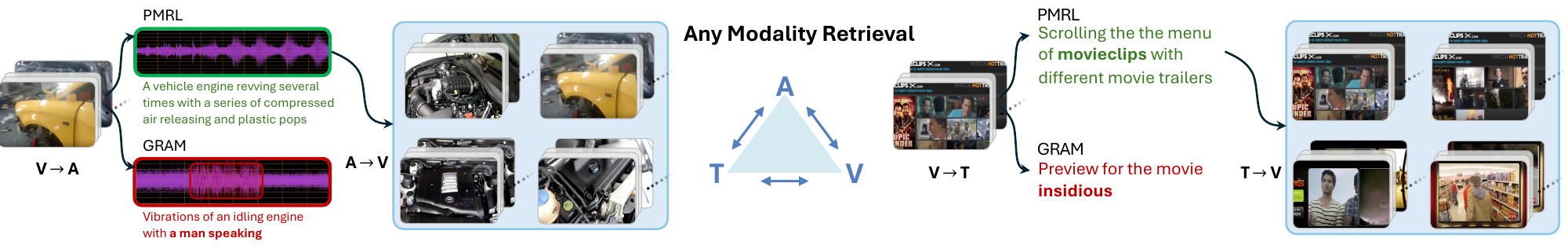}
    \caption{\textbf{The examples of any modality retrieval.} With a unified space, different modalities can retrieve others. PMRL is capable of retrieving from any modality pair with higher accuracy.} 
    \label{fig:any-modal}
\end{figure*}

\textbf{Any modality retrieval.} PMRL is capable of encouraging full alignment without a predefined anchor, making it more stable for retrieval between any modalities, exemplified by Figure~\ref{fig:any-modal}. We analyze the retrieval results among different modality pairs, \textit{e.g.}, vision-audio, compared with GRAM, as illustrated in Figure~\ref{fig:anchor-free}. 
Compared to GRAM, we achieve higher performance for any modality retrieval (colored in blue) in most cases. The retrieval performance is not only greatly improved in text-related modalities, but also in other modalities. For instance, the performance on V$\rightarrow$A retrieval boosts for all datasets, especially for AudioCaps.
Due to the limited page, we provide more detailed results in terms of Recall@5 and Recall@10 in Appendix~\ref{appendix:any_modal}.

\begin{figure}[t]
    \centering
    \includegraphics[width=\linewidth]{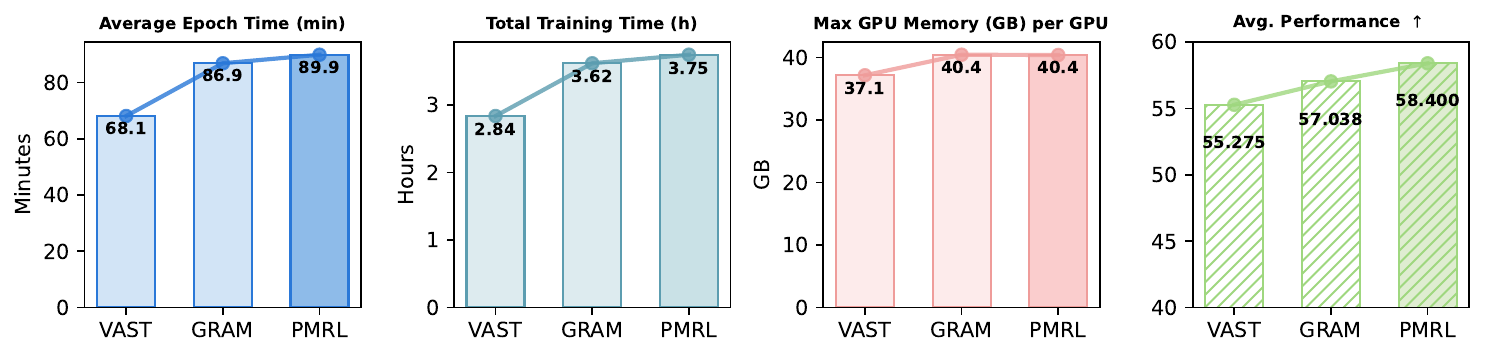}
    \vspace{-2mm}
    \caption{{Comparisons on time cost, memory usage, and average performance for three methods.}}
    \label{fig:efficiency}
\end{figure}

\begin{figure*}
    \begin{minipage}[t]{0.6\linewidth}
        \centering  
        \includegraphics[width=\linewidth]{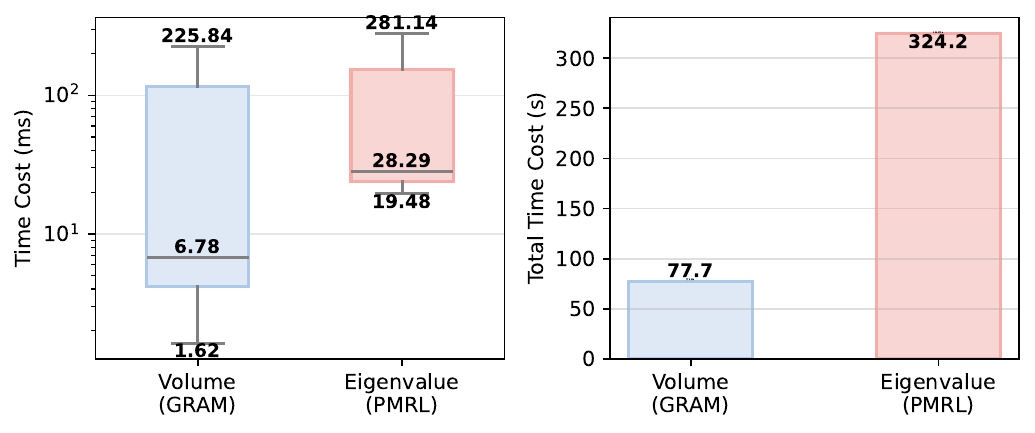}
        \caption{{Averaged time cost statistics of volume (GRAM) and eigenvalue (PMRL) computations within one single training forward step and in total.}}
        \label{fig:svd_cost}
    \end{minipage}
    \hfill
    \begin{minipage}[t]{0.34\linewidth}
        \centering  
        \includegraphics[width=0.94\linewidth]{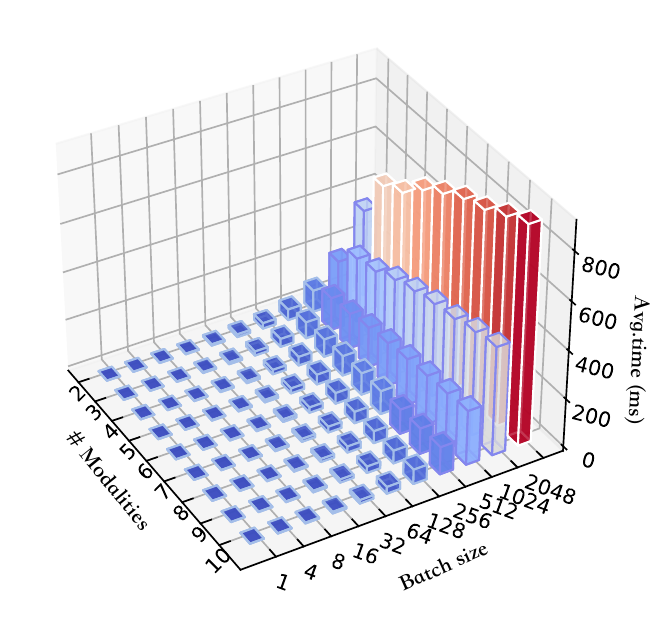}
        \caption{{Time costs of one SVD computation \textit{w.r.t.} different numbers of modalities and batch sizes.}}
        \label{fig:svd_cos_simulated}
        \end{minipage}
\end{figure*}

{\textbf{Efficiency analysis.} 
PMRL approaches full modality alignment by manipulating singular values via SVD decomposition. To assess its practical viability, we compare the time and memory costs against VAST and GRAM, as shown in Figure~\ref{fig:efficiency}. When trained in the same environment, the methods exhibit the following order of resource consumption: VAST $<$ GRAM $<$ PMRL. However, the difference between GRAM and PMRL is marginal. Given the performance gains, PMRL demonstrates strong practical viability.
We further analyze the specific computational overhead introduced by volume calculation (GRAM) and eigenvalue decomposition (PMRL) during the forward pass, and simulate SVD costs for scenarios with increased modalities and larger batch sizes (Figures~\ref{fig:svd_cost} and \ref{fig:svd_cos_simulated}). While SVD computation is relatively more expensive than volume computation for a single forward pass, the absolute cost remains low, approximately 324 s per epoch, resulting in a negligible difference over the entire training stage.
Consequently, the SVD overhead is acceptable for most scenarios. Since the number of modalities is typically small, its impact on time cost is minimal. In contrast, while increasing the batch size to 2048 causes a noticeable increase in computation time, this extreme scenario is impractical as it would incur prohibitive memory costs.
}

\begin{figure*}[t]
    \centering
    \includegraphics[width=\linewidth]{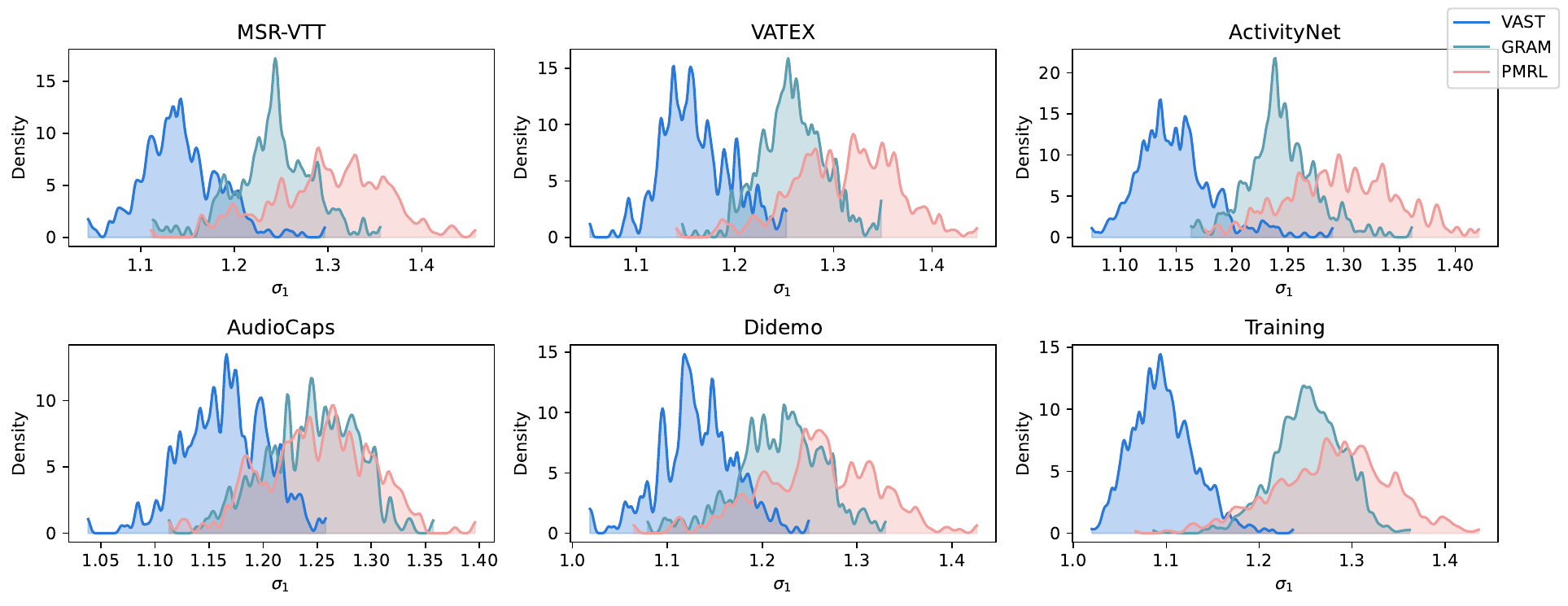}
    \caption{{Comparison of $\sigma_{1}$ distributions for three methods, indicating the trend toward rank-1 approximation in the Gram matrix of final generated features.}}
    \label{fig:sigma1}
\end{figure*}

\begin{figure*}
    \centering
    \includegraphics[width=\linewidth]{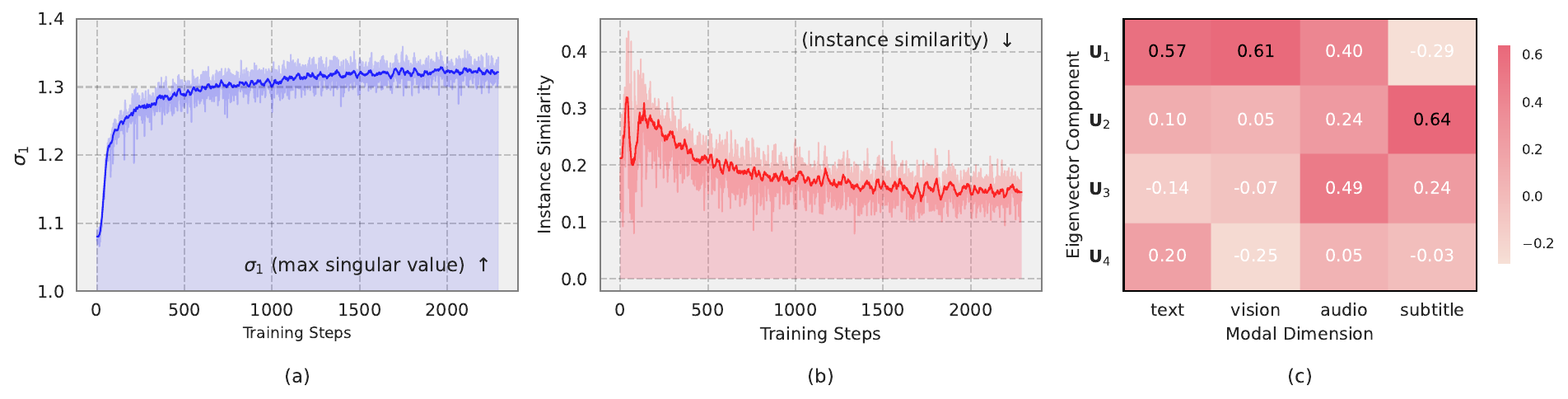}
    \caption{\textbf{Singular value analysis for PMRL.} Subfigure (a) illustrates the increase of the maximum singular value along with the training procedure induced by $\mathcal{L}^{\mathcal{M}}$. Subfigure (b) showcases the decrease of instance-wise similarity regularized by $\mathcal{L}^{\mathcal{M}'}$. Subfigure (c) depicts the contribution of each eigenvector to reconstruct the modal representation, which is interpreted by $\mathbf{V}$.}
    \label{fig:analysis}
\end{figure*}

\textbf{Maximum singular value.}
We illustrate the changes of the maximum singular values along with the training procedure in Figure~\ref{fig:analysis} (a). The result suggests an increasing trend, which can be attributed to our proposed principled learning objective. The singular value reaches a plateau afterward, indicating the convergence of the training. {$\sigma_1$ denotes the portion of shared components in GRAM matrix towards being rank-1. We further illustrate the different $\sigma_1$ distributions of $\sigma_1$ across different methods to validate the satisfaction of approaching rank-1 of the final learned features. Observed from Figure~\ref{fig:sigma1}, features learned from PMRL clearly manifest a larger $\sigma_1$ compared to other methods. This distribution plotting also corresponds to the performance order in which PMRL outperforms others, completing the empirical support for our theoretical assumption.}

\textbf{Instance-wise regularization.} We also investigate the effectiveness of principled regularization, intuited by keeping instances away, in terms of the leading eigenvector. To this end, we first measure the cosine similarity among leading eigenvectors (depicted in Figure~\ref{fig:analysis} (b)). Initially, the optimization is unstable, but continual regularization can still ensure its decrease, thereby enhancing the separability between instances. 
GRAM also implicitly introduces instance-wise regularization by comparing the volumes among instances. To isolate its impact, we modify GRAM to exclude this regularization, directly minimizing volume as $\mathcal{L} = \frac{1}{N} \sum_{i=1}^N \text{Vol}(\mathbf{Z}_i)$. Results, shown in Table~\ref{tab:no_reg}, reveal a performance drop for both PMRL models, underscoring the importance of instance-wise regularization. Note that GRAM exhibits more obvious degradation, indicating greater instability without regularization.

\textbf{Modality contribution interpretation.} 
PMRL also offers certain interpretability on modality contribution via SVD analysis, which is a core technique of our method. SVD decomposes the multimodal representation matrix $\mathbf{Z}$ into $\mathbf{U} \mathbf{\Sigma} \mathbf{V}$, where $\mathbf{U}$ represents transformed directions (eigenvectors), $\mathbf{\Sigma}$ contains singular values indicating the importance of each direction, and $\mathbf{V}$ shows how these directions are allocated to reconstruct different modalities. Therefore, the contribution of each modality to alignment can be roughly measured by $\mathbf{V}$ if we focus on the modality relevance to the first eigenvector (\textit{i.e.}, $\mathbf{U}_1$). 
To visualize this, we average the absolute values of $\mathbf{V}$ in terms of instances to create a confusion matrix, as shown in Figure~\ref{fig:analysis} (c), which highlights the relationships between modalities and the eigenvectors.
Observed from this confusion matrix, text ($m_1$) and vision ($m_2$) are strongly tied to the leading eigenvector $\mathbf{U}_1$. The audio modality also shares a large proportion with $\mathbf{U}_1$, suggesting it shares overlap with text and vision, though to a lesser extent. In contrast, subtitle modality mostly corresponds to the second eigenvector $\mathbf{U}_2$. 
These findings indicate the well-aligned bimodality, \textit{i.e.}, vision and text in semantics, also revealing the interpretability of PMRL for multimodal representation learning.

\begin{figure}
    \centering
    \includegraphics[width=0.96\linewidth]{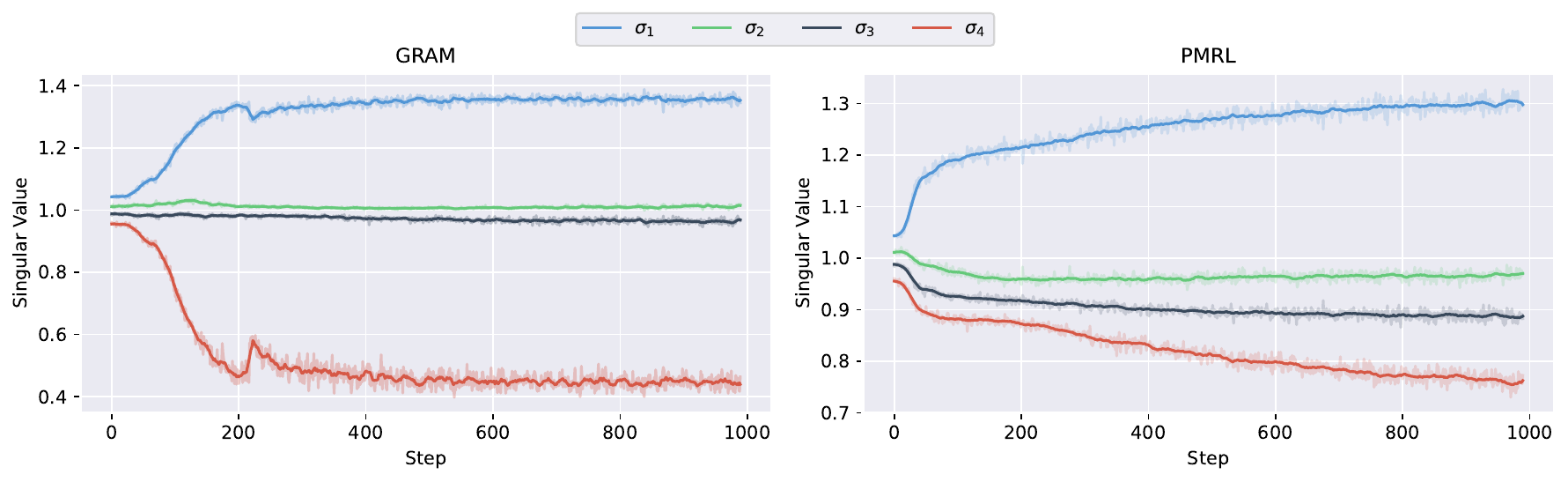}
    \caption{\textbf{The trends of singular values when training the model from scratch.} GRAM mainly focuses on minimizing one singular value, while PMRL minimizes all except the largest one simultaneously. }
    \vspace{-2mm}
    \label{fig:eigenvalues}
\end{figure}

\begin{table}
    \centering
    \caption{The performance comparison without instance-wise regularization ({w/o reg.}) \textit{w.r.t.} Recall@1 for T$\rightarrow$V.}
\label{tab:no_reg}
\resizebox{0.66\linewidth}{!}{
\begin{tabular}{l|llll} 
\toprule
   & MSR-VTT & DiDeMo & ActivityNet & VATEX \\ 
  {w/o reg.} & \small R@1 & \small R@1 & \small R@1 & \small R@1 \\ 
\midrule
 GRAM  & 50.9       &  40.2     & 20.1      &  58.7  \\           
 PMRL & \textbf{53.7 \textcolor{darkblue}{\small(+2.8)}}    & \textbf{50.2 \textcolor{darkblue}{\small(+10.0)}}  & \textbf{53.6 \textcolor{darkblue}{\small(+33.5)}}  & \textbf{80.0 \textcolor{darkblue}{\small(+21.3)}}  \\          
\bottomrule
\end{tabular}
}
\end{table}

{\textbf{Higher rank regularization.} We relax the rank regularization constraint to encourage a higher rank to validate the efficacy of PMRL's focus on the largest singular value, \textit{i.e.}, rank-1. We design a modified loss objective that maximizes the top-2 singular values, calculated as $(\sigma_{1}+ \sigma_{2})/\sum\sigma_{i}$. We denote this variant as PMRL$_{\sigma_1+\sigma_2}$ and report its performance on multimodal cross-modal retrieval tasks in Table~\ref{tab:pmrl_sigma1_and_2}. Overall, optimizing the rank of the Gram matrix for the top-2 singular values results in a slight performance drop compared to the rank-1 optimization. While higher-rank optimization preserves more distinct information from different modalities, rank-1 optimization enforces a single leading direction. This encourages different modal features to align more closely, avoiding modal-specific deviations to achieve full alignment. These results further validate the rationale behind our objective design.} 

\begin{table}[t]
\centering
\caption{{Performance comparison for video-text and audio-text retrieval across different variants of PMRL.}}
\label{tab:pmrl_sigma1_and_2}
\begin{adjustbox}{width=0.7\linewidth,center}

\begin{tabular}{l|cc|cc|cc|cc} 
\toprule
\multirow{2}{*}{} & \multicolumn{2}{c|}{ActivityNet}                     & \multicolumn{2}{c|}{VATEX}                           & \multicolumn{2}{c|}{AudioCaps}                       & \multicolumn{2}{c}{Clotho}                           \\
                        & T$\rightarrow$V                      & V$\rightarrow$T                       & T$\rightarrow$V                      & V$\rightarrow$T                       & T$\rightarrow$A                      & A$\rightarrow$T                       & T$\rightarrow$A                      & A$\rightarrow$T                       \\ 
\midrule
PMRL                    & \textbf{56.0}                     & \textbf{49.6}                      & \textbf{80.5}                     & \textbf{75.2}                      & \textbf{36.1}                     & 33.9                      & \textbf{16.8}                     & \textbf{16.1}                      \\ 
\midrule
PMRL$_{\sigma_1+\sigma_2}$                  & 55.9 & 49.5 & 79.6 & 73.6 & 34.8 & \textbf{34.9} & 16.6 & 16.0  \\
\bottomrule
\end{tabular}
\end{adjustbox}
\end{table}

\textbf{Eigenvalues analysis.}
\label{sec:eigenvalue_analysis}
Figure~\ref{fig:eigenvalues} illustrates the trends of singular values during the training of a model from scratch, comparing two methods: GRAM and PMRL. The GRAM method primarily focuses on minimizing one specific singular value, as evidenced by the significant decline of the red line ($\sigma_4$) over the training steps, while the singular values of $\sigma_2$ and $\sigma_3$ remain relatively stable. In contrast, the PMRL method minimizes all singular values except the maximum one ($\sigma_1$) simultaneously, which is reflected in the gradual decrease of $\sigma_2$, $\sigma_3$, and $\sigma_4$, while $\sigma_1$ keeps increasing. This comparison highlights the different optimization strategies employed by GRAM and PMRL, with GRAM collapsing to minimize the minimum singular value and PMRL optimizing multiple values concurrently.

\begin{table}[t]
    \centering
    \caption{The performance comparison with noise added to input and output ({w/ noise}) \textit{w.r.t.} AUC and ACC.}
    \label{tab:robust}
    \resizebox{0.64\linewidth}{!}{
    \begin{tabular}{l|ll|ll|ll} 
    \toprule
        & \multicolumn{2}{c|}{VAST} & \multicolumn{2}{c|}{GRAM} & \multicolumn{2}{c}{PMRL} \\ 
       {w/ noise} & \small AUC & \small ACC & \small AUC & \small ACC& \small AUC & \small ACC \\ 
       \midrule
     Input  &  71.5  &  64.4   &61.0    & 57.4   & \textbf{79.2 \textcolor{darkblue}{\small(+8.7)}}   &\textbf{66.2 \textcolor{darkblue}{\small(+1.8)}}   \\           
     Output &  72.9  & 66.4    & 50.0  &  57.1   &  \textbf{77.2 \textcolor{darkblue}{\small(+4.3)}} & \textbf{70.3 \textcolor{darkblue}{\small(+3.9)}}   \\          
    \bottomrule
    \end{tabular}
    }
\end{table}

\textbf{Robustness analysis.}
We show the robustness of PMRL to noise from inputs and outputs following the discussion in Section~\ref{sec:further_analysis}. 
We conduct the controlled experiments by adding Gaussian noise scaled by 0.4 to the normalized input features and randomly flipping class labels with a probability of 0.3. The results on ABIDE are reported in Table~\ref{tab:robust}. Despite performance degradation due to noise across all methods, PMRL consistently outperforms others, reflecting the robustness in principle of maximizing the maximum singular value to encourage full alignment.

\section{Conclusion and Discussion}

In this paper, we propose to strengthen the dominance of the maximum singular value about multimodal representations and distinguish the corresponding leading eigenvectors from instances to encourage full multimodal alignment. The proposed method is grounded on the theoretical insight that connects the multimodal alignment and the rank of Gram matrices. Novel learning objectives are afterward introduced to maximize the maximum singular value and regularize instance-wise separability. A series of empirical results demonstrates the effectiveness of our PMRL framework and further showcases its rational design.

Our work provides new opportunities for multimodal representation learning by reframing the full alignment problem to resolving a rank-1 approximation. The proposed novel paradigm eliminates anchor constraints, empowering the model to self-discover the leading direction for alignment adaptively. PMRL provides certain interpretability for modality contributions to alignment, and demonstrates robustness to noise. Modalities, such as MRI in the medical application, can also be handled well by PMRL with enhanced multimodal representations. 
However, in this work, we do not focus on balancing the alignment and modality distinctness, and PMRL requires the concurrency of modalities to construct a unified representation space.
Built upon our insight on approaching full alignment, future work can explore (1) trading off the perfect alignment and distinctness of multimodal representations according to our theoretical grounding, (2) scaling up training data and model parameters to develop more powerful multimodal representations, (3) incorporating emerging modalities into PMRL, and (4) exploring the generalization capability to novel modalities of PMRL with theoretical guidance and empirical support.

\clearpage
\onecolumn

\renewcommand{\cftsecfont}{\normalsize} 
\renewcommand{\cftsubsecfont}{\normalsize} 
\renewcommand{\cftbeforesecskip}{12pt}      
\renewcommand{\cftbeforesubsecskip}{12pt}

\renewcommand{\contentsname}{Appendix}
\addtocontents{toc}{\protect\setcounter{tocdepth}{2}}

  \appendix
  \tableofcontents

    \clearpage
\section{Theoretical Analysis}

\subsection{Proof of Lemma~\ref{lem:rank1}}
\label{appendix:proof_rank1}
\textit{Recall.} Let $\mathbf{G} \in \mathbb{R}^{k \times k}$ be a symmetric Gram matrix with diagonal entries equal to 1, \textit{i.e.}, $\mathbf{G}_{i,i} = 1$, and off-diagonal entries defined as $\mathbf{G}_{i,j} = \langle \mathbf{z}^i, \mathbf{z}^j \rangle$, where each $\mathbf{z}^i \in \mathbb{R}^d$ satisfies $\|\mathbf{z}^i\| = 1$. Then the following are equivalent: (1) $\mathbf{G}_{i,j} = 1$ for all $i, j$,
and (2) $\operatorname{rank}(\mathbf{G}) = 1$.

\begin{proof}

$(1) \Rightarrow (2)$:  
Suppose that $\mathbf{G}_{i,j} = 1$ for all $i, j$, \textit{i.e.},
\begin{align}
\mathbf{G} = \begin{bmatrix}
1 & 1 & \cdots & 1 \\
1 & 1 & \cdots & 1 \\
\vdots & \vdots & \ddots & \vdots \\
1 & 1 & \cdots & 1
\end{bmatrix}
= \mathbf{1} \mathbf{1}^\top,\quad \mathbf{1} = [1, 1, \dots, 1]^\top \in \mathbb{R}^k.
\end{align}
Then $\mathbf{G}$ is the outer product of a single vector with itself, so it has rank at most 1. Since $\mathbf{G} \neq \mathbf{0}$, we conclude $
\operatorname{rank}(\mathbf{G}) = 1$.
This proves the first direction.

$(2) \Rightarrow (1)$:  
Suppose that $\operatorname{rank}(\mathbf{G}) = 1$. Then $\mathbf{G}$ can be written as an outer product of two vectors:
\begin{align}
    \mathbf{G} = \mathbf{u} \mathbf{v}^\top = c \mathbf{v} \mathbf{v}^\top,
\end{align}
for some $\mathbf{u}, \mathbf{v} \in \mathbb{R}^k$. Since $\mathbf{G}$ is symmetric, we have $\mathbf{u} = c \mathbf{v}$ for some scalar $c$.
Because $\mathbf{G}$ is a Gram matrix, it is also positive semidefinite. Therefore, $c > 0$, and we can normalize $\mathbf{v}$ such that:
\begin{align}
\mathbf{G} = \mathbf{v} \mathbf{v}^\top.
\end{align}
Since $\mathbf{G}_{i,i} = \langle \mathbf{z}^i, \mathbf{z}^i \rangle = \|\mathbf{z}^i\|^2 = 1$, we have
$\mathbf{v}_i^2 = 1$, which implies $\mathbf{v}_i = \pm 1$. However, recall that $\mathbf{G}_{i,j} = \langle \mathbf{z}^i, \mathbf{z}^j \rangle = \mathbf{v}_i \mathbf{v}_j \ge 0$.
Therefore, $\mathbf{v}_i$ must all have the same sign (they are all $+1$). We then obtain:
\begin{align}
    \mathbf{v} = \mathbf{1}, \quad \Rightarrow \quad \mathbf{G} = \mathbf{1} \mathbf{1}^\top  \quad \Rightarrow \quad \mathbf{G}_{i,j} = 1, \quad \forall\ i, j.
\end{align}
\end{proof}

This equivalence captures the ideal case in multimodal alignment, where all modality representations from the same instance are perfectly aligned.

\subsection{Proof of Theorem~\ref{thm:eigenvalue}}
\label{appendix:proof_eigenvalue}

\textit{Recall.} Let $\mathbf{Z} = [\mathbf{z}^{m_1}, \dots, \mathbf{z}^{m_k}] \in \mathbb{R}^{d \times k}$ be a matrix of normalized modality representations from the same instance, \textit{i.e.}, $\|\mathbf{z}^{m_i}\| = 1$ for all $i$, and let $\sigma_1$ denote its maximum singular value. Then, (1) maximizing $\sigma_1$ maximizes the pairwise cosine similarities among $\{\mathbf{z}^m\}_{m=1}^k$, and (2) $\mathrm{rank}(\mathbf{G})=1$ is achieved if and only if $\sigma_1 = \sqrt{k}$.

\begin{proof}
According to the Eckart-Young theorem~\cite{eckart1936approximation} (see Lemma~\ref{lem:rank_svd}), the optimal rank-1 approximation $\tilde{\mathbf{Z}}$ of $\mathbf{Z}$ in the Frobenius norm is $\tilde{\mathbf{Z}} = \sigma_1 \mathbf{u}_1 \mathbf{v}_1^\top$, 
and the corresponding approximation error is:
\begin{align}
    \|\mathbf{Z} - \tilde{\mathbf{Z}}\|_F^2 = \sum_{i=2}^k \sigma_i^2.
\end{align}
Since $\|\mathbf{Z}\|_F^2 = \sum_{i=1}^k \sigma_i^2 = k$ (due to normalization $\|\mathbf{z}^i\| = 1$), we have:
\begin{align}
    \max \sigma_1 \iff \min \sum_{i=2}^k \sigma_i^2 \iff \min \|\mathbf{Z} - \tilde{\mathbf{Z}}\|_F^2.
\end{align}
Therefore, maximizing $\sigma_1$ minimizes the rank-1 approximation error. The perfect alignment can be achieved with $\sigma_1 = \sqrt{k}$.

\textit{Sufficiency:} If $\sigma_1 = \sqrt{k}$, then $\sigma_2 = \cdots = \sigma_k = 0$, meaning $\mathbf{Z}$ is exactly rank-1:
\begin{align}
  \mathbf{Z} = \sqrt{k} \, \mathbf{u}_1 \mathbf{v}_1^\top, \quad \text{with } \mathbf{v}_1 = \frac{1}{\sqrt{k}} \mathbf{1}_k.
\end{align}
This implies $\mathbf{z}^1 = \mathbf{z}^2 = \cdots = \mathbf{z}^k = \mathbf{u}_1$, achieving perfect alignment.

\textit{Necessity:} Conversely, if $\mathbf{z}^1 = \mathbf{z}^2 = \cdots = \mathbf{z}^k = \mathbf{u}_1$, then:
\begin{align}
  \mathbf{Z} = \mathbf{u}_1 \mathbf{1}_k^\top,
\end{align}
which has $\sigma_1 = \sqrt{k}$ and $\sigma_2 = \cdots = \sigma_k = 0$.
\end{proof}

The Gram matrix $\mathbf{G} = \mathbf{Z}^\top \mathbf{Z}$ has eigenvalues $\sigma_1^2 \geq \sigma_2^2 \geq \cdots \geq \sigma_k^2$. When $\sigma_1 \to \sqrt{k}$, $\mathbf{G} \to \mathbf{1}_k \mathbf{1}_k^\top$, meaning $\mathbf{z}^i{}^\top \mathbf{z}^j \to 1$ for all $i,j$. Therefore, maximizing $\sigma_1$ maximizes pairwise cosine similarities.

\subsection{Gradient Analysis}
\label{appendix:gradient}
In this section, we provide the gradient analysis for the proposed singular value-based contrastive loss:
$$
\mathcal{L}^{\mathcal{M}} = -\frac{1}{N}\sum_{i=1}^{N}\log\frac{\exp(\sigma_1/\tau)}{\sum_{j=1}^{k}\exp(\sigma_j/\tau)},
$$
where $\sigma_1$ denotes the maximum singular value of the normalized representation matrix $\mathbf{Z} \in \mathbb{R}^{d \times k}$, constructed from $k$ modality-specific embeddings of the same instance.

Let us define the softmax-normalized weights over the singular values as:
\begin{align}
    p_j = \frac{\exp(\sigma_j / \tau)}{\sum_{j'=1}^k \exp(\sigma_{j'} / \tau)} \rightarrow
    \mathcal{L}^{\mathcal{M}} = -\frac{1}{N} \sum_{i=1}^N \log p_1^{(i)},
\end{align}
where $p_1^{(i)}$ denotes the softmax weight corresponding to the maximum singular value $\sigma_1^{(i)}$ of the $i$-th instance.

\textbf{Instance-level.} 
For simplicity, we focus on one instance and drop the subscript $i$. The generalization to multiple instances follows directly. Using the chain rule and the earlier result $\frac{\partial \sigma_j}{\partial \mathbf{Z}} = \mathbf{u}_j \mathbf{v}_j^\top$, we compute:
\begin{align}
    \frac{\partial \mathcal{L}^{\mathcal{M}}}{\partial \mathbf{Z}} & = \sum_{j=1}^k \frac{\partial \mathcal{L}^{\mathcal{M}}}{\partial \sigma_j} \cdot \frac{\partial \sigma_j}{\partial \mathbf{Z}} \quad \\
    & = \sum_{j=1}^k \frac{\partial \mathcal{L}^{\mathcal{M}}}{\partial \sigma_j} \cdot \mathbf{u}_j \mathbf{v}_j^\top \quad \quad \Big(\frac{\partial \sigma_i}{\partial \mathbf{Z}} = \mathbf{u}_i \mathbf{v}_i^\top\Big)\\
    & = \frac{1}{\tau} \left[ (p_1 - 1)\mathbf{u}_1 \mathbf{v}_1^\top + \sum_{j=2}^k p_j \mathbf{u}_j \mathbf{v}_j^\top \right] .\\& \quad \quad \Big(\frac{\partial \mathcal{L}^{\mathcal{M}}}{\partial \sigma_j}
        = \frac{1}{\tau}
        \begin{cases}
        p_1 - 1, & j = 1 \\
        p_j, & j > 1
        \end{cases}\Big)
\end{align}
This expression reveals how the gradient shapes the learning dynamics: 
\begin{itemize}[leftmargin=*]
    \item The term $(p_1 - 1)\mathbf{u}_1 \mathbf{v}_1^\top$ pulls the leading direction $\mathbf{u}_1$ stronger, encouraging all columns of $\mathbf{Z}$ to align along $\mathbf{u}_1$. 
    \item The terms $p_j \mathbf{u}_j \mathbf{v}_j^\top$ for $j > 1$ act to suppress other directions, pushing the representation space into a lower-dimensional subspace aligned with $\mathbf{u}_1$.
\end{itemize}

\textbf{Modality-level.} Let us denote the $m$-th column of $\mathbf{Z}$ as $\mathbf{z}^m$, representing the embedding of the $m$-th modality. Then, the gradient of the loss with respect to $\mathbf{z}^m$ can be extracted from the above expression:
\begin{align}
    \frac{\partial \mathcal{L}^{\mathcal{M}}}{\partial \mathbf{z}^m} = \frac{1}{\tau} \sum_{j=1}^k \frac{\partial \mathcal{L}^{\mathcal{M}}}{\partial \sigma_j} \cdot \mathbf{u}_j \mathbf{v}_{jm},
\end{align}
where $\mathbf{v}_{jm}$ is the $m$-th entry of the right singular vector $\mathbf{v}_j$. This implies that each modality’s representation is updated proportionally to its projection onto the dominant singular direction $\mathbf{u}_1$, weighted by the softmax probability $p_j$.

\section{Implementation Details}

\subsection{Training and Benchmark Datasets}
\label{appendix:datasets}
We employ the training dataset \textbf{VAST-150K}~\cite{cicchetti2024gramian}, which is sampled from VAST-27M~\cite{chen2023vast}, following the training setting of GRAM~\cite{cicchetti2024gramian}. 
VAST-27M is sampled from the large-scale HD\_VILA\_100M corpus~\cite{xue2022hdvila}, involving diverse categories of music, gaming, education, entertainment, animals, and more. Four modalities, \textit{i.e.}, video, audio, caption, and subtitle, are collected for each example.
More than that, we adopt several benchmark datasets as follows:

\begin{itemize}
    \item \textbf{MSR-VTT}~\cite{chen2011collecting} is a large-scale video description dataset comprising approximately 10,000 short video clips (10-20 seconds each) sourced from YouTube, totaling around 200,000 video-text pairs. Each clip is annotated with 20 human-generated English captions, covering diverse scenarios such as sports, music, and daily activities. In our experiment, we extract the audio, which serves as one of three modalities. We adopt the standard split. 
    \item \textbf{DiDeMo}~\cite{anne2017localizing} focuses on localized video descriptions, containing about 10,000 videos sourced from Flickr. Each video is annotated with four textual descriptions tied to specific temporal segments, emphasizing semantic diversity and temporal localization. These four short sentences are concatenated and arranged in temporal order. The official split is used.
    \item \textbf{ActivityNet}~\cite{krishna2017dense} is a large-scale video dataset tailored for human activity recognition, comprising approximately 20,000 YouTube videos totaling around 648 hours. It covers 200 activity classes (\textit{e.g.}, cooking and sports) with temporally annotated segments and associated descriptions. Approximately 3,000 videos are unavailable online. Therefore, we remove them for our evaluation with the adopted official split. 
    \item \textbf{VATEX}~\cite{wang2019vatex} is a multilingual video description dataset containing about 41,000 10-second video clips derived from the Kinetics-600 dataset, which covers 600 human activity categories. There are also some unavailable videos online. We adopt the split following~\cite{rohrbach2017movie, chen2023vast} and exclude these examples for evaluation.
    \item \textbf{AudioCaps}~\cite{kim2019audiocaps} is a large-scale audio description dataset, featuring approximately 51,000 10-second audio clips sourced from AudioSet. Each clip is paired with 1-5 human-annotated English captions describing diverse sound events (\textit{e.g.}, natural, human, or mechanical sounds). We evaluate text-audio retrieval, following the same split protocol by~\cite{oncescu2021audio}.
    \item \textbf{Clotho}~\cite{drossos2020clotho} contains 6,974 (in its expanded version) audio clips (15-30 seconds each) sourced from Freesound, each annotated with 5 detailed English captions. By emphasizing complex and diverse sound scenes, Clotho provides rich semantic descriptions for audio events. Its official split is adopted.
    \item \textbf{ABIDE}~\cite{craddock2013neuro} is a neuroimaging dataset comprising brain imaging data (sMRI, fMRI, and \textit{etc.}) from 871 subjects, including individuals with autism spectrum disorder (ASD) and healthy controls. Collected from multiple international sites, it includes functional connectivity data, structural imaging, and metadata (\textit{e.g.}, age and gender). We utilize the metadata to construct the textual attribute as a modality. We follow the split protocol proposed by~\cite{kan2022brain}. Note that we do not employ the cross-validation method for evaluation. 
    \item {\textbf{VGGSound}~\cite{chen2020vggsound} is a large-scale audio-visual dataset designed for sound recognitionand audio-vision correspondence with labels. Collected from YouTube, it contains over 200,000 video clips covering a wide range of "in-the-wild" sounding objects. We use its downsized version, with 2,000 samples in the test split for zero-shot audio classification. }
    \item {\textbf{UCF101}~\cite{soomro2012ucf101} is a foundational dataset for video action recognition with videos collected from YouTube. It features 101 distinct action categories ranging from sports to playing musical instruments and general body movements. During the testing, we select 10 examples for each category to balance the distribution. }
    \item {\textbf{ImageNet}~\cite{deng2009imagenet} is a massive image database organized according to the WordNet hierarchy, instrumental in advancing deep learning for computer vision. We select the version with 1,000 object classes for the evaluation. }
    \item {\textbf{NYUDv2}~\cite{SilbermanNYUDv2} is a premier dataset for indoor scene understanding, capturing paired RGB and depth information using a Microsoft Kinect. We use the preprocessed version of the original NYU Depth V2 datasets. Due to the loss of the test labels, we use this dataset for cross-modal retrieval. Wherein, 47,584 examples are used for training, and 654 examples are for testing. }
    \item {\textbf{TVL}~\cite{fu2024touch} is a multimodal dataset designed to align tactile (touch) sensations with visual and linguistic data. We use the provided split for training and testing. }
    \item {\textbf{Houston13}~\cite{debes2014hyperspectral} is a specialized remote sensing dataset originally distributed for the 2013 IEEE GRSS Data Fusion Contest. It uses two distinct modalities: Hyperspectral Imagery (HSI) with 144 spectral bands and LiDAR. We take its land cover classes as an additional modality for multimodal alignment.}
\end{itemize}

\subsection{Baselines}
\label{appendix:baselines}

We briefly introduce the used baselines in multimodal learning and autism classification. 

\begin{itemize}
\item \textbf{Frozen} \cite{bain2021frozen} is an end-to-end trainable model adapting ViT and Timesformer architectures with spatio-temporal attention, trained on both large-scale image and video captioning datasets using a curriculum learning approach. 
\item \textbf{UMT} \cite{liu2022umt} is the first to jointly optimize moment retrieval and highlight detection in videos by integrating multi-modal (visual-audio) learning, treating moment retrieval as keypoint detection with a query generator and decoder.
\item \textbf{UMT-L} \cite{li2023unmasked} enhances data efficiency by masking low-semantics video tokens and selectively aligning unmasked tokens with an image foundation model as an unmasked teacher, enabling faster convergence and multimodal compatibility.
\item \textbf{OmniVL} \cite{wang2022omnivl} introduces a unified transformer-based foundation model that supports both image-language and video-language tasks through a single architecture, utilizing decoupled joint pretraining to enhance spatial and temporal vision-language modeling. 
\item \textbf{TVTSv2} \cite{zeng2023tvtsv2} proposes a degradation-free pre-training strategy for video foundation models, preserving the text encoder's generalization by freezing shallow layers and tuning deep layers, while using a transcript sorting task with masking for scalable training.
\item \textbf{CLIP4Clip} \cite{luo2022clip4clip} adapts a CLIP image-language pre-training model for end-to-end video-text retrieval. 
\item \textbf{ViCLIP} \cite{wang2023internvid} is a video-text representation learning model based on ViT-L, trained on a large-scale video-centric multimodal dataset with over 7 million videos and 234M clips, paired with 4.1B words of detailed descriptions. 
\item \textbf{VideoCoCa} \cite{yan2022videococa} adapts a pretrained image-text contrastive captioner model for video-text tasks by leveraging its generative and contrastive attentional pooling layers for flattened frame embeddings.
\item \textbf{Norton} \cite{lin2023multi} employs video-paragraph and clip-caption contrastive losses for video-language learning, which filters irrelevant clips and captions, realigns asynchronous pairs, and uses a soft-maximum operator to handle fine-grained frame-word misalignments. 
\item \textbf{ImageBind} \cite{girdhar2023imagebind} introduces a joint embedding method across six modalities with image-paired data, leveraging large-scale vision-language models to extend zero-shot capabilities to new modalities.
\item \textbf{InternVideo-L} \cite{wang2022internvideo} presents a general video foundation model that combines generative masked video modeling and discriminative video-language contrastive learning to pretrain video representations.
\item \textbf{HiTeA} \cite{ye2023hitea} introduces a hierarchical temporal-aware video-language pre-training framework with cross-modal moment exploration to model detailed video moment representations and multi-modal temporal relation exploration to capture temporal dependencies across video-text pairs at varying time resolutions. 
\item \textbf{mPLUG-2} \cite{xu2023mplug} introduces a modularized multi-modal pretraining framework with a multi-module composition network, sharing universal modules for modality collaboration while disentangling modality-specific modules to address entanglement. 
\item \textbf{VALOR-L} \cite{liu2024valor} proposes an end-to-end pretraining framework that jointly models vision, audio, and language using three separate encoders for modality-specific representations and a decoder for multimodal conditional text generation. 
\item \textbf{TEFAL} \cite{ibrahimi2023audio} introduces a text-conditioned feature alignment method for text-to-video retrieval, utilizing two independent cross-modal attention blocks to align text queries with audio and video representations separately.
\item \textbf{Bimodal T2M} \cite{arora2024text} proposes a hierarchical multimodal video retrieval model that enhances text-to-video retrieval by creating a shared embedding space using task-specific contrastive loss functions, designed to maximize mutual information between textual and cross-modal representations.
\item \textbf{T-MASS} \cite{wang2024text} introduces a stochastic text modeling approach for text-video retrieval, representing text as a flexible, resilient semantic ``text mass'' through a similarity-aware radius module and supporting text regularization. 
\item \textbf{vid-TLDR} \cite{choi2024vid} proposes a training-free token merging method for video Transformers, enhancing efficiency by merging background tokens using a saliency-aware strategy that leverages attention maps to focus on salient regions and drop irrelevant background tokens.
\item \textbf{VideoPrism-b} \cite{zhao2024videoprism} introduces a general-purpose video encoder pretrained on a diverse corpus of 36M video-caption pairs and 582M clips with noisy text, using a global-local distillation and token shuffling approach to enhance masked autoencoding. 
\item \textbf{LanguageBind} \cite{zhu2023languagebind} proposes a multi-modal pretraining framework that extends video-language pretraining to multiple modalities by using a frozen language encoder from VL pretraining as the semantic bind.
\item \textbf{AVFIC}~\cite{nagrani2022learning} propose a multimodal transformer-based model trained on a new large-scale, weakly labeled audio-video captioning dataset with millions of paired clips and captions without additional manual effort. 
\item \textbf{VIP-ANT}~\cite{zhao2021connecting} leverages shared image modality as a pivot in a tri-modal embedding space for audio-text alignment, eliminating the need for parallel audio-text data. 
\item \textbf{VAST} \cite{chen2023vast} trains a multimodal foundation model on the VAST-27M dataset, which is created by integrating vision and audio captions generated by separately trained captioners with subtitles using a large language model.
\item \textbf{GRAM} \cite{cicchetti2024gramian} aligns multiple modalities in a higher-dimensional embedding space using a contrastive loss function that minimizes the Gramian volume of the $k$-dimensional parallelotope spanned by modality vectors.
\end{itemize}

For VAST, we utilize its pre-trained base model for zero-shot prediction. Due to it not releasing the fine-tuned versions, we fine-tune it for downstream tasks to evaluate its performance under the fine-tuning setting. 
For GRAM, we directly use their well-trained model weights for evaluation for two settings. 
Below, we introduce the baselines in autism classification. 

\begin{itemize}
\item \textbf{AE-FCN}~\cite{rakic2020improving} integrates functional connectivity patterns from fMRI and volumetric correspondences of gray matter from sMRI, using a combination of unsupervised stacked autoencoders and supervised multilayer perceptrons.
\item \textbf{GCN}~\cite{parisot2018disease} utilizes graph convolutional networks by representing populations as a sparse graph, where nodes incorporate imaging-based feature vectors and edges integrate phenotypic information as weights.
\item \textbf{BrainNetCNN}~\cite{kawahara2017brainnetcnn} employs a convolutional neural network (CNN) to predict neurodevelopmental outcomes. It uses several convolutional filters (edge-to-edge, edge-to-node, node-to-graph) with the topological locality from structural brain networks. 
\item \textbf{DGM}~\cite{kazi2022differentiable} introduces a learnable function that predicts edge probabilities in graphs, enabling end-to-end training with convolutional graph neural network layers to infer graph structures directly from data. 
\item \textbf{BrainNetTF}~\cite{kan2022brain} models brain networks as graphs with fixed-size, ordered nodes using connection profiles as node features for natural positional information and learns pairwise ROI connection strengths via efficient attention weights.
\item \textbf{VanillaTF} is a simplified version of BrainNetTF, which consists of a two-layer Transformer and a concat-based readout.
\end{itemize}

VAST and GRAM are also utilized for comparison by equipping BrainNetTF, MLP, and BERT as modal encoders. PMRL follows the same model architecture for a fair comparison.

\subsection{Evaluation Metrics}
\label{appendix:metrics}
We evaluate the multimodal retrieval tasks with \textbf{Recall} as the metric, and evaluate autism classification (in binary) by using \textbf{AUC} (Area Under the Curve) and \textbf{ACC} (Accuracy) as metrics. Recall@$K$ measures the proportion of relevant items successfully retrieved within the top $K$ results. 
Let $\{q_1, q_2, \dots, q_N\} $ denote the set of queries, and for each query $ q_i $, let $ \mathcal{R}_i^K \subseteq \mathcal{D} $ denote the set of top $ K $ retrieved items from the dataset $ \mathcal{D} $. Let $ \mathcal{S}_i \subseteq \mathcal{D} $ denote the set of true positive (relevant) items associated with query $ q_i $. 
Recall at $ K $ is defined as:
$
\text{Recall}@K = \frac{1}{N} \sum_{i=1}^{N} \frac{|\mathcal{R}_i^K \cap \mathcal{S}_i|}{|\mathcal{S}_i|}.
$
In cases where each query corresponds to exactly one correct match, this simplifies to the ratio of queries for which the correct item appears among the top $K$ retrieved results.
Accuracy is defined as
$\text{Accuracy} = \frac{1}{n} \sum_{i=1}^{n} \mathbb{I}(\hat{y}_i = y_i)$,
where $ n $ is the total number of samples and $ \mathbb{I}(\cdot) $ is the indicator function.
Let $ f(\mathrm{x}_i) \in [0, 1] $ denote the model’s predicted probability for the sample $\mathrm{x}_i$. The AUC estimates the probability that a randomly chosen positive sample is ranked higher than a randomly chosen negative one:
$
\text{AUC} = \frac{1}{n_+ n_-} \sum_{i: y_i=1} \sum_{j: y_j=0} \mathbb{I}(f(\mathrm{x}_i) > f(\mathrm{x}_j))
$,
where $ n_+ $ and $ n_- $ are the numbers of positive and negative samples, respectively.

\subsection{Hyperparameter Settings}
\label{appendix:hyperparameters}

We utilize AdamW~\cite{loshchilov2017decoupled} as the optimizer, where the learning rate is set to 2$\times10^{-5}$, $\beta_1=0.9$, and $\beta_2=0.98$. The linear schedule is employed for warmup, with a warmup ratio of 0.1. The weight decay is 0.01, and the gradient norm is limited to 2. 
All the representations are transformed into 512 dimensions. For our PMRL model, $\tau_1$ is set to 0.05, $\tau_2$ is set to 0.1; $\lambda_1$ is configured as 1.0 and $\lambda_2$ is 0.1. For autism classification, we employ 5-times trials and report the averaged performance. We set the learning rate to 1$\times 10^{-4}$ and Adam~\cite{kinga2015method} as the optimizer. The output representations are transformed into 128 dimensions. We adjust $\tau_2$ to 0.4, and other settings are kept the same.

\subsection{Model Architecture}
\label{appendix:model_architecure}
We design the PMRL model architecture following the well-developed VAST model. Specifically, the vision encoder is set to use EVAClip-ViT-G~\cite{sun2023eva}, with 1.3B parameters. The input resolution for visual data is configured to 224$\times$224 pixels. The text encoder is implemented with BERT, with the maximum caption length limited to 40 and the subtitle length to 70.
The audio encoder is configured to use the BEATs model~\cite{chen2023beats}. The audio input is processed into 64 mel-frequency bins, and the target input length is set to 1,024 frames.
{We also implement PMRL with another model architecture of ImageBind to involve more modalities and diverse tasks. Specifically, ImageBind includes vision, audio, text, depth, thermal, and IMU modalities. We add an additional projector after the backbone to achieve an efficient fine-tuning on the VAST-150K dataset. We directly use the vision to process the tactile modality on the TVL dataset to demonstrate the adaptation on novel modality. We implement VAST with multiple contrastive losses to align different modalities. All the settings on the model architecture are the same for VAST, GRAM, and PMRL to ensure a fair comparison and validation. }

{For multimodal hyperspectral imaging tasks, we use a convolutional neural network as the projector for LiDAR and HSI, followed by a mean pooling to obtain the final representations. The dimension is set as 256. The text label is viewed as a special modality for alignment and is processed as learnable embeddings. The features of LiDAR and HSI are averaged for the classification evaluation. }

For multimodal neuron imaging tasks, we implement the PMRL model by equipping it with an fMRI encoder as BrainNetTF~\cite{kan2022brain} (built upon a graph transformer model), an sMRI encoder as a 2-layer MLP, and a text encoder as BERT as well. 
Resting-state fMRI data is preprocessed via a CPAC pipeline and a specified brain parcellation atlas (\textit{i.e.}, CC200). For each subject, the mean time series of each brain region was extracted using the selected atlas. Subsequently, two types of functional connectivity matrices were computed: Pearson correlation and partial correlation matrices, representing the pairwise relationships between brain regions. sMRI features are extracted from FreeSurfer-processed outputs. ComBat harmonization is applied to the sMRI features to mitigate site and batch effects, using site, age, sex, IQ, and diagnostic label as covariates. The resulting sMRI features are concatenated into a matrix. For the textual features, we combine age and gender attributes as ``age: $<$attr\_age$>$, gender: $<$attr\_gender$>$'' for each subject. The multimodal representations are averaged and fed to a 3-layer MLP that returns the predictions in binary for classification. We replace $\mathcal{L}_{\text{IM}}$ with the classification loss in implementing VAST, GRAM, and PMRL.

\subsection{Pseudo Code}
\label{appendix:pseudocode}
To facilitate reproducibility, we additionally provide the pseudocode of PMRL. These materials demonstrate the straightforward implementation of integrating PMRL in just a few steps.

\begin{tcolorbox}[colframe=gray!10, colback=gray!10, coltitle=black, arc=0pt, title=\textbf{Integrating PMRL with four steps},left=0pt, right=0pt]
\begin{lstlisting}[language=Python, belowskip=0mm,]
# 1. Singular Value Decomposition on Multimodal Representations >>>
U, S, _ = torch.linalg.svd(
    torch.stack([feat_t,feat_v,feat_a,feat_s], dim=-1)
    )

# 2. Principled learning via maximum singular values >>>
loss1 = F.cross_entropy(S/self.tau1, torch.zeros(S.shape[0]).to(S.device).long())
# Implemented by cross-entropy, and the singular value at the first position is the maximum one

# 3. Principled regularization via eigenvector corresponding to the maximum singular values >>>
U1 = U[:, :, 0]
loss2 = F.cross_entropy((U1 @ U1.T)/self.tau2, torch.arange(U1.shape[0]).to(U1.device).long())

......

# 4. Combine the loss >>>
loss = loss1 + self.lambda1 * loss2 + self.lambda2 * loss_IM

\end{lstlisting}
\end{tcolorbox}

\section{Additional Results}

We provide the full results on multimodal text-video retrieval, especially in terms of Recall@1, Recall@5, and Recall@10 metrics as shown in Tables~\ref{tab:vt_zero_full_1},~\ref{tab:vt_zero_full_2},~\ref{tab:vt_finetune_full_1}, and~\ref{tab:vt_finetune_full_2}. Moreover, we illustrate more results of any modality retrieval on Recall@5 and Recall@10 in Figures~\ref{fig:anchor-free_R5} and~\ref{fig:anchor-free_R10}. We also exhibit the trends of each singular value during training to reveal the collapse of GRAM compared to PMRL, as shown in Figure~\ref{fig:eigenvalues}.

\subsection{Multimodal Text-Video Retrieval}
\label{appendix:tv_full}
We report the available results in metrics of Recall@1, Recall@5, and Recall@10 for text-video retrieval. The performances under zero-shot and fine-tuning settings are shown in Tables~\ref{tab:vt_zero_full_1}, ~\ref{tab:vt_zero_full_2} and Tables~~\ref{tab:vt_finetune_full_1},~\ref{tab:vt_finetune_full_2}, respectively. Aligning with the results reported in the main content, our PMRL method also outperforms other methods in most cases.

\begin{table*}[t]
\centering
\caption{Multimodal text-to-video (T$\rightarrow$V) and video-to-text (V$\rightarrow$T) retrieval results on zero-shot setting (\%) across MSR-VTT and DiDeMo.}
\label{tab:vt_zero_full_1}
\begin{adjustbox}{width=\textwidth,center}
\begin{tabular}{@{}l|lll|lll|lll|lll@{}}
\toprule
 & \multicolumn{6}{c|}{MSR-VTT} & \multicolumn{6}{c}{DiDeMo}   \\  & \multicolumn{3}{c|}{T$\rightarrow$V}  & \multicolumn{3}{c|}{V$\rightarrow$T}  & \multicolumn{3}{c|}{T$\rightarrow$V}  & \multicolumn{3}{c}{V$\rightarrow$T}     \\
    & R@1 &R@5 & R@10    & R@1 &R@5 & R@10  & R@1 &R@5 & R@10  & R@1 &R@5 & R@10    \\ \midrule 
Fronzen \cite{bain2021frozen} & 18.7 & 39.5 & 51.6  &  -  &  - & - & 21.1    &  46.0 & 59.2  &  - & -  &  -  \\
UMT \cite{liu2022umt} & 33.3  &  - & 66.7  &  -  &  - & - & 34.0  &  - & 68.7  &  - & -  &  - \\
UMT-L \cite{li2023unmasked}  & 40.7  & 63.4 & 71.8  & 37.1   &  - & -   & 48.6    &  72.9 & 79.0   & 49.9   &  - & - \\
OmniVL \cite{wang2022omnivl} & 42.0    &  63.0 & 73.0 &  40.7  &  - & - & 40.6  &  64.6 & 74.3   & 24.9  &  - & -\\
TVTSv2 \cite{zeng2023tvtsv2}  & 38.2  &  62.4 & 73.2      &    -   &  - & - & 34.6   &  61.9 & 71.5    &  - & - & -\\
ViCLIP \cite{wang2023internvid}  & 42.4  &  - & - & 41.3   &  - & -    & 18.4   &  - & -    & 27.9   &  - & -   \\
VideoCoCa \cite{yan2022videococa}   & 34.3  &  57.8 & 67.0   & 64.7    &  85.2 & 67.0 & -    &  - & -      & -   &  - & -   \\
Norton \cite{lin2023multi}                    &   10.7 &  24.1 & 31.6   &  - & -&    &  - & - &     -    &  - & -  \\ 
ImageBind \cite{girdhar2023imagebind}  & 36.8   & 61.8 & 70.0  &  -   &  - & -& -     &  - & -    & -   &  - & - \\
InternVideo-L \cite{wang2022internvideo}  & 40.7  &  - & -&    39.6   &  - & - & 31.5   &  - & -   & 33.5   &  - & -    \\
HiTeA \cite{ye2023hitea}  & 34.4 &  60.0 & 69.9 &    -  &  - & -  & 43.2  &  69.3 & 79.0    & -     &  - & - \\
mPLUG-2 \cite{xu2023mplug}  & 47.1  &  69.7 & 79.0 &    -  &  - & -  & 45.7    &  71.1 & 71.1  & -    &  - & - \\
VideoPrism-b \cite{zhao2024videoprism}  & 51.4  &  - & -    & 50.2   &  - & - & -    &  - & - & -   &  - & -\\
LanguageBind \cite{zhu2023languagebind}  & 44.8    &  70.0 & 78.7  & 40.9   &  66.4 & 75.7 & 39.9   &  66.1 & 74.6  & 39.8  &  67.8 & 76.2 \\ \midrule 
VAST \cite{chen2023vast}   & 50.5   & 69.0 & 74.3 & 48.8  & 69.9 & 75.6 & 46.4 & 67.5 & 73.5 & 45.3 & 68.7 & 75.4  \\
GRAM \cite{cicchetti2024gramian} & 51.5   & 71.5 & 77.9  & 51.5  & 73.5 & 79.5  & 49.8  & 71.0 & 76.3 & 48.5  & 70.1 & 75.5 \\

\midrule
\textbf{PMRL (Ours)}  & 54.5   & 73.2 & 80.4 & 52.4   &  73.8 & 79.8 & 50.6    &  72.7 & 77.4  &  48.4 &  70.8 & 78.3 \\
 \bottomrule
\end{tabular}
\end{adjustbox}
\end{table*}

\begin{table*}[t]
\centering
\caption{Multimodal text-to-video (T$\rightarrow$V) and video-to-text (V$\rightarrow$T) retrieval results on zero-shot setting (\%) across ActivityNet and VATEX.}
\label{tab:vt_zero_full_2}
\begin{adjustbox}{width=\textwidth,center}
\begin{tabular}{@{}l|lll|lll|lll|lll@{}}
\toprule
 & \multicolumn{6}{c|}{ActivityNet} & \multicolumn{6}{c}{VATEX}   \\  & \multicolumn{3}{c|}{T$\rightarrow$V}  & \multicolumn{3}{c|}{V$\rightarrow$T}  & \multicolumn{3}{c|}{T$\rightarrow$V}  & \multicolumn{3}{c}{V$\rightarrow$T}     \\
    & R@1 &R@5 & R@10    & R@1 &R@5 & R@10  & R@1 &R@5 & R@10  & R@1 &R@5 & R@10    \\ \midrule 
UMT \cite{li2023unmasked}& 31.9   &  - & 72.0 & -  &  - & -& - &  - & -& -  &  - & - \\
UMT-L \cite{li2023unmasked}& 41.9   &  - & -& 39.4  &  - & -& - &  - & -& -  &  - & - \\
ViCLIP \cite{wang2023internvid}  & 15.1     &  - & -& 24.0  &  - & -& -  &  - & -& -  &  - & -\\
VideoCoCa \cite{yan2022videococa}   & 34.5      &  63.2 & 76.6& 33.0  &  61.6 & 75.3 & 53.2  &  83.3 & 90.1& 73.6  &  93.2 & 97.2\\
InternVideo-L \cite{wang2022internvideo}  & 30.7       &  - & -& 31.4  &  - & -& 49.5  &  - & -& 69.5  &  - & -\\
VideoPrism-b \cite{zhao2024videoprism}  & 49.6       &  - & -&  47.9  &  - & -& 62.5  &  - & -& 77.1 &  - & -\\
LanguageBind \cite{zhu2023languagebind}  & 41.0   &  68.4 & 80.0 &  39.1  &  69.8 & 81.1& -  &  - & -& -  &  - & -\\ \midrule 
VAST \cite{chen2023vast}  & 51.7  & 75.7 & 83.4 & 48.8 & 74.8 & 81.9 & 75.9 & 93.3 & 94.8 & 74.8  & 93.5 & 95.6\\
GRAM \cite{cicchetti2024gramian}  & 54.5  & 78.3 & 85.2 & 48.3  &  74.2 & 82.6 & 77.5   &  94.8 & 96.2 & 74.7  & 93.5 & 95.5\\
\midrule
\textbf{PMRL (Ours)}  & 56.0 &  80.0 & 87.4 & 49.6  &  76.0 & 85.6 &80.5 &  95.4 & 96.4  & 75.2  &  93.8 & 95.5\\ \bottomrule
\end{tabular}
\end{adjustbox}
\end{table*}

\begin{table*}[t]
\centering
\caption{Multimodal text-to-video (T$\rightarrow$V) and video-to-text (V$\rightarrow$T) retrieval results on finetuning setting (\%) across MSR-VTT and DiDeMo.}
\label{tab:vt_finetune_full_1}
\begin{adjustbox}{width=\textwidth, center}
\begin{tabular}{@{}l|lll|lll|lll|lll@{}}
\toprule
 & \multicolumn{6}{c|}{MSR-VTT} & \multicolumn{6}{c}{DiDeMo}   \\  & \multicolumn{3}{c|}{T$\rightarrow$V}  & \multicolumn{3}{c|}{V$\rightarrow$T}  & \multicolumn{3}{c|}{T$\rightarrow$V}  & \multicolumn{3}{c}{V$\rightarrow$T}     \\
    & R@1 &R@5 & R@10    & R@1 &R@5 & R@10  & R@1 &R@5 & R@10  & R@1 &R@5 & R@10    \\ \midrule 
UMT-L \cite{li2023unmasked} & 58.8$^*$& 81.0$^*$    & 87.1$^*$    & 58.6$^*$& -   & -      & 70.4$^*$&  90.1$^*$    & 93.5$^*$   & 65.7$^*$ & -   & -    \\
CLIP4Clip \cite{luo2022clip4clip} & 44.5  & 71.4   & 81.6   & 45.9 & -   & -    & 43.4 & 70.2   & 80.6  & 43.6  & -   & -     \\
ViCLIP \cite{wang2023internvid}  & 52.5  & -   & -  & 51.8 & -   & -       & 49.4   & -   & -     & 50.2   & -   & -   \\
InternVideo-L \cite{wang2022internvideo}  & 55.2$^*$ & -   & -  &    57.9$^*$  & -   & -   & 57.9$^*$  & -   & -     & 59.1$^*$   & -   & -   \\
HiTeA \cite{ye2023hitea}  & 46.8 & 71.2   & 81.9  &  -   & -   & -   & 56.5    & 81.7   & 89.7  & - & -   & -   \\
mPLUG-2 \cite{xu2023mplug}  & 53.1 & 77.6   & 84.7  &    -   & -   & -  & 56.4   & 79.1   & 85.2    & -  & -   & -   \\
VALOR-L \cite{liu2024valor}  & 54.4 & 79.8   & 87.6  &    -  & -   & -   & 57.6   & 83.3   & 88.8    & -   & -   & -   \\
TEFAL \cite{ibrahimi2023audio}  & 52.0 & 76.6   & 86.1 &    -  & -   & -   & -  & -   & -     & -    & -   & -   \\
Bimodal T2M \cite{arora2024text}  & 36.8 & -   & -  &    -  & -   & -   & -  & -   & -     & -    & -   & -  \\
T-MASS \cite{wang2024text}  & 52.7 & 77.1   & 85.6  &    -  & -   & -   & 53.3  & 80.1   & 87.7     & -  & -   & -    \\
vid-TLDR \cite{choi2024vid}  & 58.5$^*$ & 81.3$^*$   & 86.9$^*$   &  - & -   & -   & 70.4$^*$ & 90.5$^*$   & 94.0$^*$  & - & -   & -  \\
\midrule 
VAST \cite{chen2023vast}  & 64.4 & 84.3 & 90.4   &  64.3  & 86.2 & 92.9 & 68.4  & 86.9 & 90.1   & 65.4 & 88.0 & 90.7 \\
GRAM \cite{cicchetti2024gramian} & 60.0 & 79.6 & 84.3   & 61.8 & 80.9 & 85.2  & 68.7 & 86.0 & 89.2   & 65.7 & 86.8 & 91.2 \\
\midrule
\textbf{PMRL (Ours)}  & 61.2 & 80.4   & 85.5    & 60.7& 82.2   & 86.4    & 70.2 & 87.5   & 91.0    &  66.4 & 87.8  & 90.9 \\
 \bottomrule
\end{tabular}
\end{adjustbox}
\end{table*}

\begin{table*}[t]
\centering
\caption{Multimodal text-to-video (T$\rightarrow$V) and video-to-text (V$\rightarrow$T) retrieval results on finetuning setting (\%) across ActivityNet and VATEX. }
\label{tab:vt_finetune_full_2}
\begin{adjustbox}{width=\textwidth,center}
\begin{tabular}{@{}l|lll|lll|lll|lll@{}}
\toprule
 & \multicolumn{6}{c|}{ActivityNet} & \multicolumn{6}{c}{VATEX}   \\  & \multicolumn{3}{c|}{T$\rightarrow$V}  & \multicolumn{3}{c|}{V$\rightarrow$T}  & \multicolumn{3}{c|}{T$\rightarrow$V}  & \multicolumn{3}{c}{V$\rightarrow$T}     \\
    & R@1 &R@5 & R@10    & R@1 &R@5 & R@10  & R@1 &R@5 & R@10  & R@1 &R@5 & R@10    \\ \midrule 
UMT-L \cite{li2023unmasked}   & 66.8$^*$  & 89.1$^*$    & 94.9$^*$    & 64.4$^*$  & -   & -   & 72.0$^*$ & -   & -   & 86.0$^*$ & -   & -   \\
CLIP4Clip \cite{luo2022clip4clip} & 40.5  & 72.4  & -   & 41.6  & -   & -   & 55.9 & 89.2 & 95.0  & 78.3 & -   & -   \\
ViCLIP \cite{wang2023internvid}  & 49.8   & -   & -    & 48.1 & -   & -   & - & -   & -   & - & -   & -   \\
InternVideo-L \cite{wang2022internvideo}  & 62.2$^*$      & -   & -   & 62.8$^*$ & -   & -   & 71.1$^*$ & -   & -   & 87.2$^*$ & -   & -   \\

VALOR-L \cite{liu2024valor} & 63.4      & 87.8   & 94.1   & -  & -   & -   & 76.9& 96.7   & 98.6   &-& -   & -   \\
TEFAL \cite{ibrahimi2023audio} & -      & -   & -   & -  & -   & -   & 61.0 & 90.4 & 95.3   &- & -   & -   \\
T-MASS \cite{wang2024text}  & -      & -   & -   & -  & -   & -   & 65.6 & 93.9   & 97.2   &-& -   & -   \\
vid-TLDR \cite{choi2024vid} & 65.2$^*$ &  88.7$^*$   & 94.5$^*$   & -  & -   & -   & - & -   & -   & - & -   & -    \\
\midrule 
VAST \cite{chen2023vast} & 68.1  & 89.5   & 95.7   &  65.4 & 88.7  & 94.9    & 83.1   & 98.1& 99.2   &  81.3  &98.4&99.6  \\

GRAM \cite{cicchetti2024gramian} & 67.6 & 89.4 & 95.4    & 65.0 & 88.4 & 94.5  & 82.5 & 98.0 & 98.9   & 80.6 & 98.0 & 99.2  \\
\midrule
\textbf{PMRL (Ours)}  &   68.2 & 89.1  & 94.6 & 66.4 & 88.4   & 94.1   &84.1 & 97.3   & 98.3   & 83.2 & 97.8  &  98.4  \\
 \bottomrule
\end{tabular}
\end{adjustbox}
\end{table*}

\subsection{Any Modality Retrieval}
\label{appendix:any_modal}
\begin{figure*}
    \centering
    \includegraphics[width=0.98\linewidth]{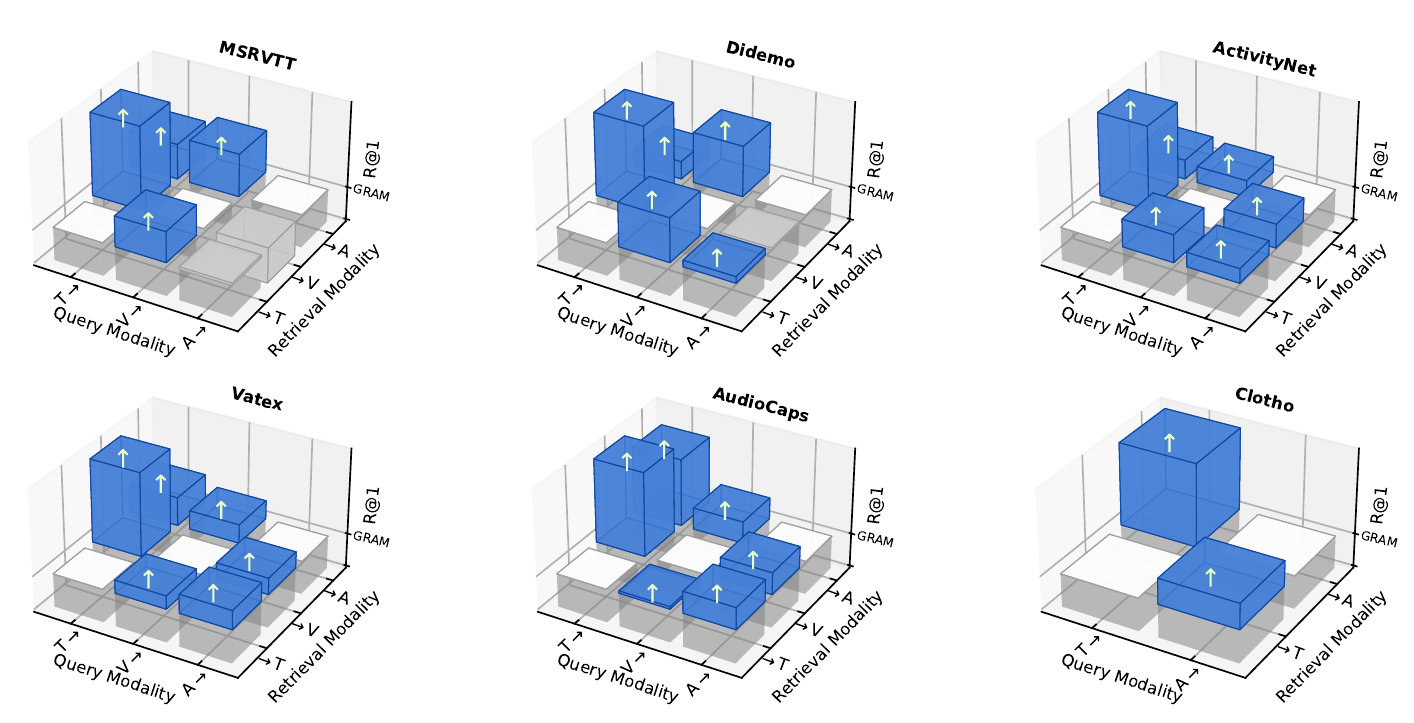}
    \caption{\textbf{Performance comparison of PMRL \textit{v.s.} GRAM in terms of Recall@5 for any modality retrieval across 6 benchmark datasets.} Blue regions highlight where PMRL outperforms GRAM, while gray regions indicate the opposite. Diagonal regions (colored in white) represent self-modal retrieval, which is not meaningful for comparison.}
    \label{fig:anchor-free_R5}
\end{figure*}

\begin{figure*}
    \centering
    \includegraphics[width=0.98\linewidth]{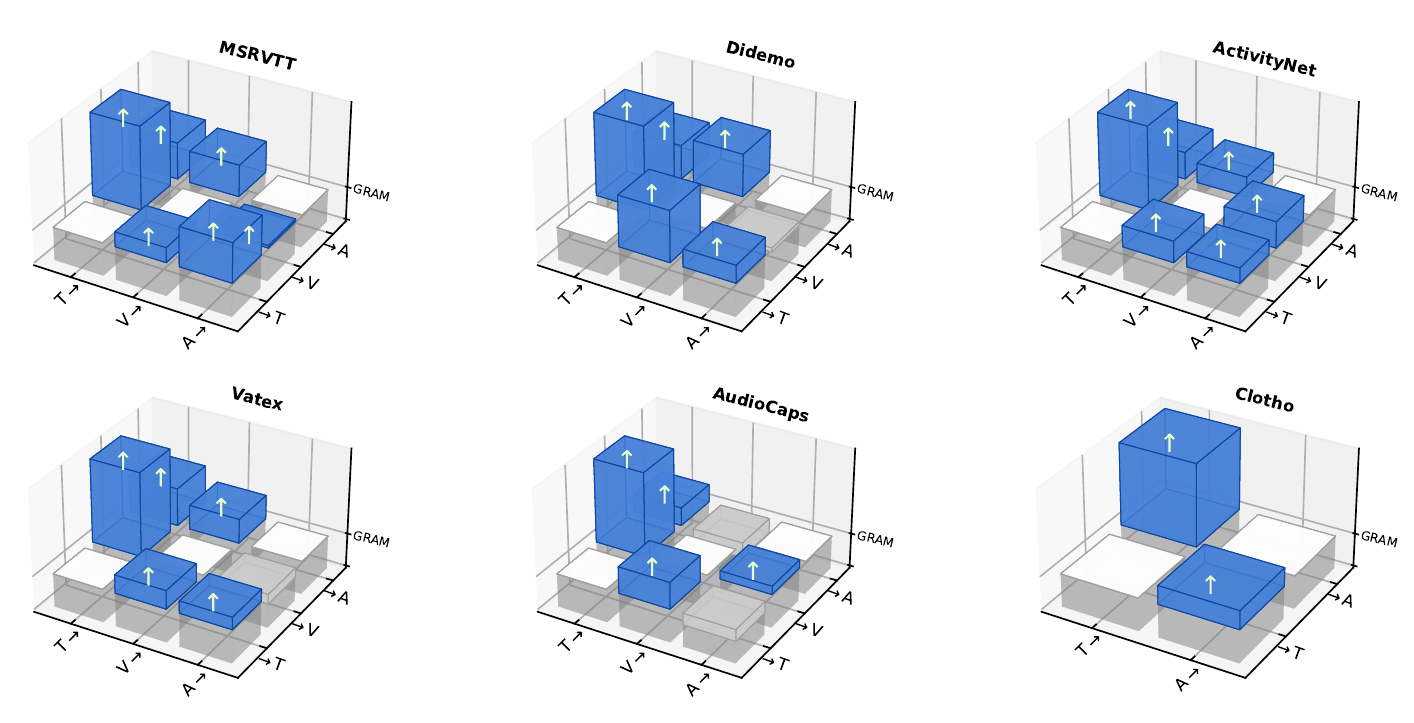}
    \caption{\textbf{Performance comparison of PMRL \textit{v.s.} GRAM in terms of Recall@10 for any modality retrieval across 6 benchmark datasets.} Blue regions highlight where PMRL outperforms GRAM, while gray regions indicate the opposite. Diagonal regions (colored in white) represent self-modal retrieval, which is not meaningful for comparison.}
    \label{fig:anchor-free_R10}
\end{figure*}

Figures~\ref{fig:anchor-free_R5} and~\ref{fig:anchor-free_R10} illustrate a performance comparison between PMRL and GRAM across six benchmark datasets (MSR-VTT, Didemo, ActivityNet, Vatex, AudioCaps, and Clotho) in terms of Recall@5 and Recall@10 for any-modality retrieval. Blue regions highlight areas where PMRL outperforms GRAM, indicating superior retrieval accuracy, while gray regions show where GRAM performs better. Diagonal regions, colored in white, represent self-modal retrieval, which is not meaningful for comparison and thus excluded from the analysis. The 3D bar charts visualize the performance differences across various modalities (denoted as A, T, V, \textit{etc.}), with the height of the bars reflecting the recall scores, providing a clear visual representation of the relative strengths of PMRL and GRAM across different datasets and retrieval conditions. From the observations, it can be concluded that PMRL generally outperforms GRAM, with the strongest performance observed in text-relevant modality retrieval. Additionally, PMRL demonstrates significant improvement over GRAM in non-text-relevant modality retrieval as well, like V$\rightarrow$T.

\section{Reproducibility}
\label{sec:repeoducibility}
We provide implementation details, involving illustrative algorithm descriptions and pseudo-code in Appendix~\ref{appendix:pseudocode}. The source code will be publicly released for reproducibility.

\section{Limitations}
\label{sec:limitations}
PMRL advances multimodal alignment by optimizing the maximum singular value of Gram matrices and ensuring instance-wise separability. However, the resource constraints, like the updated YouTube policies on video downloads, prevented us from collecting large-scale, high-quality multimodal datasets needed to fully enhance PMRL’s capabilities. Therefore, PMRL has to employ continual training on pre-trained models. Despite the limitation, experimental results demonstrate the effectiveness of PMRL and the rationale of our core design.

\clearpage
\bibliography{reference}
\bibliographystyle{unsrt}

\end{document}